\documentclass[11pt]{article}
\usepackage[margin=1in]{geometry}
\usepackage{times}
\usepackage{natbib}
\setcitestyle{authoryear,round,citesep={;},aysep={,},yysep={;}}
\usepackage{xcolor}

\usepackage{amsmath,amsfonts,bm}

\def\eqref#1{equation~\ref{#1}}
\def\1{\bm{1}}

\DeclareMathAlphabet{\mathsfit}{\encodingdefault}{\sfdefault}{m}{sl}
\SetMathAlphabet{\mathsfit}{bold}{\encodingdefault}{\sfdefault}{bx}{n}

\usepackage{hyperref}
\usepackage{url}

\usepackage{graphicx}
\usepackage{tikz}
\usetikzlibrary{arrows.meta,calc,positioning}
\usepackage{booktabs}
\usepackage{multirow}
\usepackage{array}
\usepackage{enumitem}
\usepackage{tabularx}
\usepackage[normalem]{ulem}
\newcounter{example}
\usepackage{capt-of}
\usepackage[utf8]{inputenc}

\usepackage{listings}
\usepackage{newunicodechar}

\newunicodechar{←}{$\leftarrow$}
\newunicodechar{ℤ}{$\mathbb{Z}$}
\newunicodechar{ℝ}{$\mathbb{R}$}
\newunicodechar{∣}{$\mid$}
\newunicodechar{≠}{$\neq$}
\newunicodechar{”}{''}
\newunicodechar{“}{``}
\newunicodechar{’}{'}
\newunicodechar{—}{---}
\newunicodechar{–}{--}
\newunicodechar{₄}{$_4$}
\newunicodechar{₃}{$_3$}
\newunicodechar{₂}{$_2$}
\newunicodechar{σ}{$\sigma$}
\newunicodechar{≃}{$\simeq$}
\newunicodechar{⊢}{$\vdash$}
\newunicodechar{×}{$\times$}
\newunicodechar{≤}{$\le$}
\newunicodechar{∑}{$\sum$}
\newunicodechar{→}{$\to$}

\newcommand{\benchname}{MathAdv}
\usepackage{CJKutf8}

\usepackage{qcircuit,braket}

\newcommand{\cfM}{\mathfrak{M}}

\title{MathAdv: What Theorem Provers Know, Reason, Formalize, and Generalize}

\newcommand{\authorentry}[3]{%
  \shortstack{\small #1%
    \if\relax\detokenize{#3}\relax\else\textsuperscript{#3}\fi
    \if\relax\detokenize{#2}\relax\else\\[-0.2em]{\scriptsize\texttt{#2}}\fi}}
\author{\begin{tabular}{@{}ccc@{}}
\authorentry{Jiaxin Yuan}{jyuan98@umd.edu}{1} &
\authorentry{Udari Madhushani Sehwag}{udari.madhu.703@gmail.com}{6} &
\authorentry{Connor Martinez Lockhart}{connorl@umd.edu}{1} \\[0.8em]
\authorentry{Xiaoyu Liu}{xiaoyu.liu1231@gmail.com}{1} &
\authorentry{Shayan Shabihi}{shabihi@umd.edu}{1,6} &
\authorentry{Jiaqi Wang}{jwang3737@gatech.edu}{3} \\[0.8em]
\authorentry{Chenghao Deng}{dengch16@umd.edu}{1} &
\authorentry{Xiayimei Han}{xhan1115@umd.edu}{1} &
\authorentry{Vlassis Mastrantonis}{vm429@cornell.edu}{4} \\[0.8em]
\authorentry{Dmitrii Gudin}{dgudin@umd.edu}{1} &
\authorentry{Shaopeng Zhu}{szhu@terpmail.umd.edu}{2} &
\authorentry{Abdirisak Mohamed}{amoham70@umd.edu}{1} \\[0.8em]
\authorentry{Bilal Aytekin}{baytekin@umd.edu}{1} &
\authorentry{Jiewen Lang}{jiewenlang@gmail.com}{2} &
\authorentry{Zezheng Song}{zsong2019@gmail.com}{1} \\[0.8em]
\authorentry{Furong Huang}{furongh@umd.edu}{1,5} &
 & \\
\multicolumn{3}{c}{\scriptsize
\textsuperscript{1}University of Maryland, College Park \quad
\textsuperscript{2}Independent Researcher \quad
\textsuperscript{3}Georgia Institute of Technology} \\[0.1em]
\multicolumn{3}{c}{\scriptsize
\textsuperscript{4}Cornell University \quad
\textsuperscript{5}All Purpose AI \quad
\textsuperscript{6}Scale AI}
\end{tabular}}
\date{}

\begin{document}

\newcommand{\xl}[1]{\textcolor{blue}{\textbf{[Xiaoyu: #1]}}}
\newcommand{\fhcomment}[1]{\textcolor{blue}{\textbf{FH: #1}}}

\definecolor{connoredit}{rgb}{0.0, 0.45, 0.45}
\newcommand{\cn}[1]{\textcolor{connoredit}{#1}}

\maketitle
\begin{abstract}
Formal theorem proving enables machine-verifiable evaluation of mathematical reasoning, yet existing benchmarks often emphasize aggregate proof accuracy, concentrate on a narrow range of mathematics, and provide limited evidence of robustness to equivalent reformulations. We introduce \benchname, a diagnostic benchmark spanning 13 domains across undergraduate- and graduate-level mathematics. Alongside Lean 4 theorem proving, \benchname\ provides up to three auxiliary tasks: multiple-choice questions that probe mathematical knowledge, fill-in-the-blank problems that isolate informal reasoning, and expert-crafted transformations that test robustness to problem presentation. 
% We construct the benchmark through domain-expert curation and an LLM-assisted, verifier-in-the-loop formalization pipeline that combines compiler feedback with independent semantic review. 
Our evaluation of contemporary theorem provers yields four findings: formalization remains a major bottleneck; performance varies substantially across mathematical domains; natural-language guidance helps general-purpose LLMs but can hinder proof-specialized models; and mathematically equivalent reformulations expose substantial robustness limitations. Together, these results show how component-wise evaluation can reveal model capabilities and failure modes that aggregate theorem-proving accuracy obscures. The dataset and evaluation scripts are available at \url{https://github.com/margotyjx/MathAdv.git}. 
% \fhcomment{Jiaxin: Please replace this public GitHub URL with an anonymized link before submission.}
\end{abstract}

% \vspace{1em}
\begin{center}
    \includegraphics[width=0.9\linewidth]{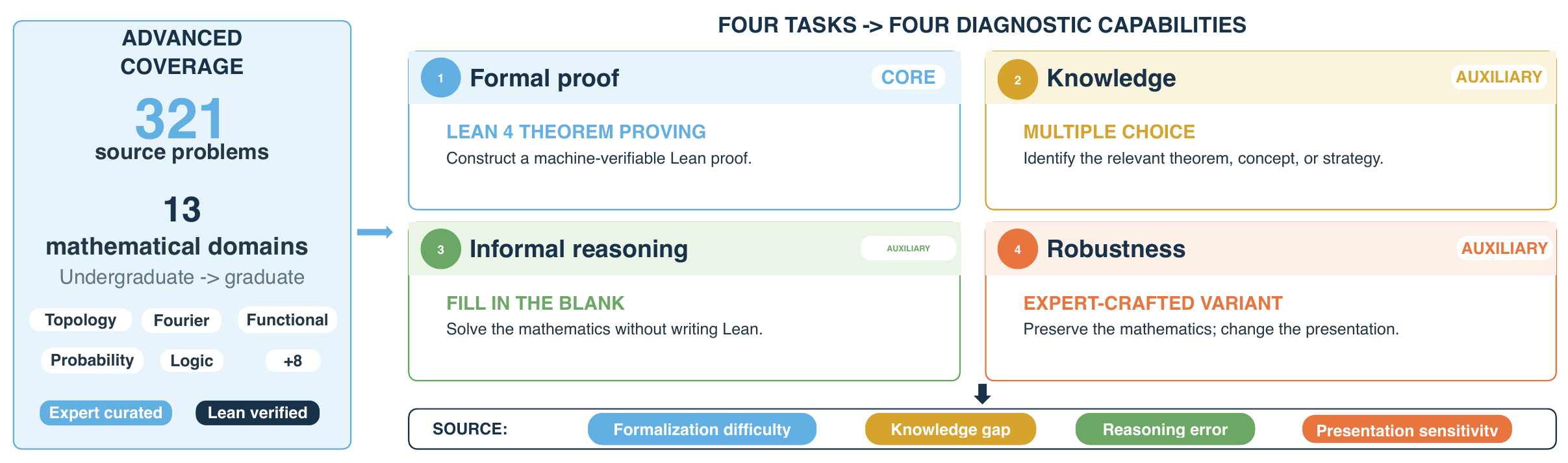}
    \captionof{figure}{\benchname\ at a glance. 
    % The benchmark combines broad coverage of advanced mathematics with four complementary task views that separately probe formal proof construction, mathematical knowledge, informal reasoning, and robustness to equivalent reformulations. This design enables component-wise diagnosis beyond aggregate theorem-proving accuracy.
    }
    \label{fig:mathadv_teaser}
\end{center}

\section{Introduction}

Mathematical reasoning is a fundamental benchmark for assessing AI because it requires abstract understanding, logical deduction, and multi-step reasoning beyond memorization~\cite{hendrycks2021math, mirzadeh2025gsmsymbolic}. It is also central to scientific discovery and other domains requiring reliable, verifiable conclusions, making its rigorous evaluation increasingly important~\cite{lewkowycz2022solvingquantitativereasoningproblems, riosgarcia2026ai}. Early benchmarks primarily assessed informal natural-language solutions by their final answers, providing limited guarantees about the validity of intermediate reasoning~\cite{lewkowycz2022solvingquantitativereasoningproblems, luo2025wizardmathempoweringmathematicalreasoning}. Formal theorem proving offers a more rigorous alternative: models construct machine-verifiable proofs in assistants such as Lean~\cite{lean4}, Coq~\cite{bertot2004coq}, and Isabelle~\cite{nipkow2002isabelle}, enabling automatic verification and precise feedback on invalid proof attempts~\cite{li2024surveydeeplearningtheorem, hubert2026olympiad}. Recent advances in language models have further strengthened proof-generation systems and intensified interest in formal reasoning~\cite{lin2025goedelproverv2scalingformaltheorem, ren2025deepseekproverv2advancingformalmathematical}.

% While recent formal mathematics benchmarks have substantially advanced theorem-proving evaluation, understanding the reasoning capabilities underlying model performance remains an important open challenge. {\color{red} Not informative, need to change.}

However, verifying a proof and understanding a model's mathematical capabilities are different evaluation goals. A proof score records success on formal statements, but leaves three questions unresolved. \textbf{First, what capabilities underlie success or failure?} Theorem proving draws on mathematical knowledge, logical deduction, and formal proof construction. Aggregate accuracy alone cannot distinguish a knowledge gap from a reasoning error or difficulty expressing an argument in Lean. \textbf{Second, across which mathematical domains do these capabilities hold?} Existing benchmarks have largely focused on competition-level problems drawn from high-school and undergraduate mathematics, with an emphasis on algebra and number theory~\cite{liu2023fimo, tsoukalas2024putnambenchevaluatingneuraltheoremprovers}, leaving broader advanced mathematics less explored. \textbf{Third, does success persist when the formulation changes?} Evaluations generally use a single, fixed formulation, although language-model reasoning can be sensitive to minor changes in problem presentation~\cite{mirzadeh2025gsmsymbolic}. Concerns about data contamination and memorization further motivate testing beyond familiar formulations~\cite{gardner-etal-2020-evaluating, Kaushik2020Learning, sakaguchi2020winogrande}. Together, these questions call for evaluation beyond proof accuracy: probing the capabilities behind a score, their domain coverage, and their robustness.
% Together, these issues make it increasingly important to determine whether successful proofs reflect robust mathematical reasoning or reliance on familiar formulations and superficial cues.
% Together, these observations motivate evaluations that move beyond theorem-proving accuracy alone and provide a more fine-grained analysis of the capabilities underlying mathematical reasoning.

We introduce \benchname, a benchmark designed to investigate these three questions within a unified evaluation framework. \textbf{First, paired tasks probe the capabilities behind proof performance.} Alongside Lean theorem proving, multiple-choice questions probe mathematical knowledge and direct-answer questions assess informal reasoning without requiring Lean proof construction. Comparing these task outcomes helps identify potential knowledge gaps, reasoning errors, and formal proof construction difficulties. \textbf{Second, broad coverage reveals domain-specific strengths and limitations.} The benchmark contains 321 undergraduate- and graduate-level problems across 13 domains, including topology, Fourier analysis, and functional analysis. Of these, 298 have Lean 4 statements; the remaining 23 are retained for auxiliary evaluations pending adequate Mathlib support. \textbf{Third, equivalent reformulations test the stability of proof success.} Expert-crafted variants preserve the underlying mathematical content while substantially changing its presentation. Each suitable problem has up to three auxiliary evaluations, connecting capability-level analysis with domain coverage and robustness testing.

Constructing \benchname\ poses two main technical challenges: designing paired tasks that probe complementary capabilities while preserving their connection to the original mathematics, and producing Lean statements that are both type-correct and semantically faithful despite uneven Mathlib coverage. We address the first through domain-expert curation and independent review: experts select nonredundant problems, construct the auxiliary tasks, and verify that transformed variants retain the reasoning required by the originals. We address the second with an LLM-assisted, verifier-in-the-loop formalization pipeline that combines compiler checks with independent semantic review.
An LLM drafts each Lean statement, compiler feedback guides iterative correction, an independent LLM and a domain expert check semantic fidelity, and a second expert performs final review. 
% For concepts lacking adequate Mathlib support, experts assess viable encodings and defer formalization when a faithful translation would require prohibitive library development.

Our evaluations show why these complementary views matter: proof performance alone obscures gaps between mathematical reasoning and formal proof construction, as well as sensitivity to domain and presentation. \textbf{First, a promising mathematical approach does not guarantee a valid Lean proof.} Models often identify a useful proof direction but fail to formalize it successfully. Reasoning hints improve general-purpose LLMs but can degrade proof-specialized theorem provers, further showing that informal guidance does not translate uniformly into proof success. \textbf{Second, mathematical competence is uneven across domains.} Performance varies substantially by subject, so aggregate accuracy conceals domain-specific strengths and weaknesses. \textbf{Third, proof success is sensitive to equivalent reformulations.} Most models solve original problems substantially more often than their expert-crafted variants, limiting the extent to which success on one presentation establishes robustness. These findings support evaluating the capabilities and conditions behind proof success, rather than treating a single proof score as a complete account of mathematical reasoning.

% Our contributions are as follows:
% \begin{enumerate}[
% itemsep=0pt,
% topsep=0pt,
% parsep=0pt,
% partopsep=0pt,
% leftmargin=1.5em,
% labelsep=0.4em
% ]

% \item \textbf{A broad, expert-reviewed benchmark for advanced formal mathematics.}
% We introduce \benchname, comprising 321 problems across 13 undergraduate- and graduate-level domains, including underrepresented areas such as topology, Fourier analysis, and functional analysis. Among them, 298 have Lean 4 statements constructed through an LLM-assisted, verifier-in-the-loop pipeline with independent expert review to ensure compilability and semantic fidelity.

% \item \textbf{A component-wise diagnostic framework for theorem proving.}
% Beyond aggregate proof accuracy, \benchname\ evaluates four capabilities: mathematical knowledge, informal reasoning, formal proof construction, and robustness to reformulation. The original Lean tasks are paired with up to three auxiliary evaluations---multiple-choice questions, fill-in-the-blank problems, and expert-crafted transformed variants---to help identify the source of model successes and failures.

% \item \textbf{A systematic characterization of contemporary theorem provers.}
% Our evaluation shows that formalization remains a major bottleneck, performance varies substantially across mathematical domains, the effectiveness of natural-language guidance depends on model specialization, and robustness to mathematically equivalent reformulations remains limited.

% \end{enumerate}

Our contributions are as follows:
\begin{enumerate}[
itemsep=0pt,
topsep=0pt,
parsep=0pt,
partopsep=0pt,
leftmargin=1.5em,
labelsep=0.4em
]
\item \textbf{An evaluation framework beyond proof accuracy:}
paired tasks that probe mathematical knowledge, informal reasoning, and formal proof construction, with equivalent reformulations to test robustness.

\item \textbf{An expert-reviewed benchmark spanning advanced mathematics:}
321 problems across 13 domains, including 298 Lean 4 statements checked for compilability and semantic fidelity, enabling domain-level analysis of these capabilities.

\item \textbf{A systematic evaluation of contemporary theorem provers:}
evidence that promising mathematical approaches can fail at formal proof construction, performance is uneven across domains, and proof success is sensitive to equivalent reformulations and model-dependent use of guidance.
\end{enumerate}
\section{Related work}

\textbf{Formal mathematics reasoning.}
Formal mathematics encodes statements and proofs within a logical system implemented by a proof assistant such as Lean~\cite{lean4}. Mathematical objects and definitions must be built from concepts recognized by the system, and every derivation must follow its logical rules. The proof assistant therefore verify each step and certify that the conclusion follows from the stated assumptions. 
% For models, this requires translating mathematical ideas into precise definitions, statements, and logically valid proof steps.

\textbf{Existing benchmark datasets.} % miniF2F, ProofNet, FormalNumina,Lean-workbook. FormalMath, Herald, IndiMathBench
Recent benchmarks evaluate formal mathematical reasoning in large language models across competition mathematics, informal-to-formal translation, formal theorem collections etc., but typically focus on specific sources, domains, or capabilities ~\cite{zheng2021minif2f,azerbayev2023proofnet,liu2023fimo,tsoukalas2024putnambenchevaluatingneuraltheoremprovers,yu2025formalmath,liu2026numina,biyani2025indimathbench}. In contrast, \benchname\ spans 13 domains in undergraduate- and graduate-level mathematics and includes direct-answer, multiple-choice, and transformed questions for diagnosis beyond theorem-proving accuracy alone. Appendix~\ref{app:benchmark_review} provides a detailed comparison.

\textbf{Formal theorem provers}
Recent work has developed a wide range of large language model approaches for both informal mathematical reasoning and formal theorem proving. Formal theorem provers range from direct proof generators to systems that use search, verifier feedback, large formal training corpora, or multiple reasoning agents~\cite{polu2020generative,azerbayev2023llemma,ying2024internlmmath,lin2024lean,shen2025real,lin2025goedelproverfrontiermodelopensource,xin2024deepseekproverv15harnessingproofassistant,wang2024theoremllama,gao2024herald,wang2025malot}. These approaches differ primarily in how they generate candidate proofs and recover from errors; Appendix~\ref{app:model_review} provides a detailed review.

% \subsection{Formal theorem benchmarks}
% What is formal theorem, Lean3-4, representative benchmark datasets. How they formalize from natural language questions

% \textbf{Previously related work:} FormalMath, Herald, IndiMathBench (geometry)

% \subsection{Formal mathematics reasoning}
% Mathematical reasoning models, formal mathematical reasoning models. 

% LLM-based models. Best-first search, tree search, lean-STAR sampling. Single-pass generation methods. Advent of language models directly in solving mathematical reasoning questions.

% \textbf{Models: }
% ReProver, Llemma, Lean-STAR, InternLM, Deepseek, TheoremLlamma, GPT-f, Deepseek-Prover, Goedel-Prover, MA-LoT. etc. 
\section{\benchname: advanced benchmarking of formal math reasoning}

% \xl{I would reformat this paragraph as in this section, we do 1 2 3 4... . for example, in this section, we first introduce the A in section \ref.., then we explain how the diagnostic questions are developed and .... this is still a bit vague.  for example, you mention Lean 4 formalization ability, but why we should care about it ? this is not explained clearly either here or before. }
% To construct a broadly representative diagnostic benchmark, human experts select undergraduate- and graduate-level problems spanning 13 mathematical domains. For each problem, they manually develop auxiliary diagnostic questions designed to distinguish domain knowledge, natural-language reasoning, Lean 4 formalization ability, and robustness to problem transformations. The corresponding Lean 4 statements are then produced through a human-in-the-loop autoformalization pipeline.
% This section describes the construction of \benchname\ and the pipeline used to formalize its problems in Lean 4.

Section~\ref{sec:benchmark_design} introduces \benchname\ and describes three complementary diagnostic question types that test natural-language problem solving, knowledge of relevant theorems and proof strategies, and robustness to equivalent reformulations. These diagnostic tasks allow us to disentangle these component abilities from formal proof construction, which requires combining them to produce a precise, verifiable proof in Lean.
In Section~\ref{sec:autoformalization_pipeline}, we present the human-in-the-loop pipeline used to translate problems into Lean 4. 

\subsection{\benchname}
\label{sec:benchmark_design}
% \benchname\ consists of 321 mathematical problems spanning 13 domains: number theory, linear algebra, abstract algebra, calculus, real analysis, complex analysis, Fourier analysis, functional analysis, probability, topology, geometry, combinatorics, and logic. The problems in \benchname\ are drawn from undergraduate- and graduate-level mathematics textbooks or contributed by domain experts. They are selected to provide broad coverage while minimizing redundancy and to expose meaningful failure modes in contemporary LLMs. Figure~\ref{fig:dist_by_category_n_example} shows the distribution of problems across domains. 
% \xl{ With these topics, we developed three types of problems: (1)Fill-in-the-blank (2) Multiple-choice  (3)Transformed dual as shown in the Figure ~/refxxxx
% }

\begin{figure}[h]
\centering
\begin{minipage}[t]{0.33\linewidth}
\vspace{0pt}
\centering
\includegraphics[width=0.8\linewidth]{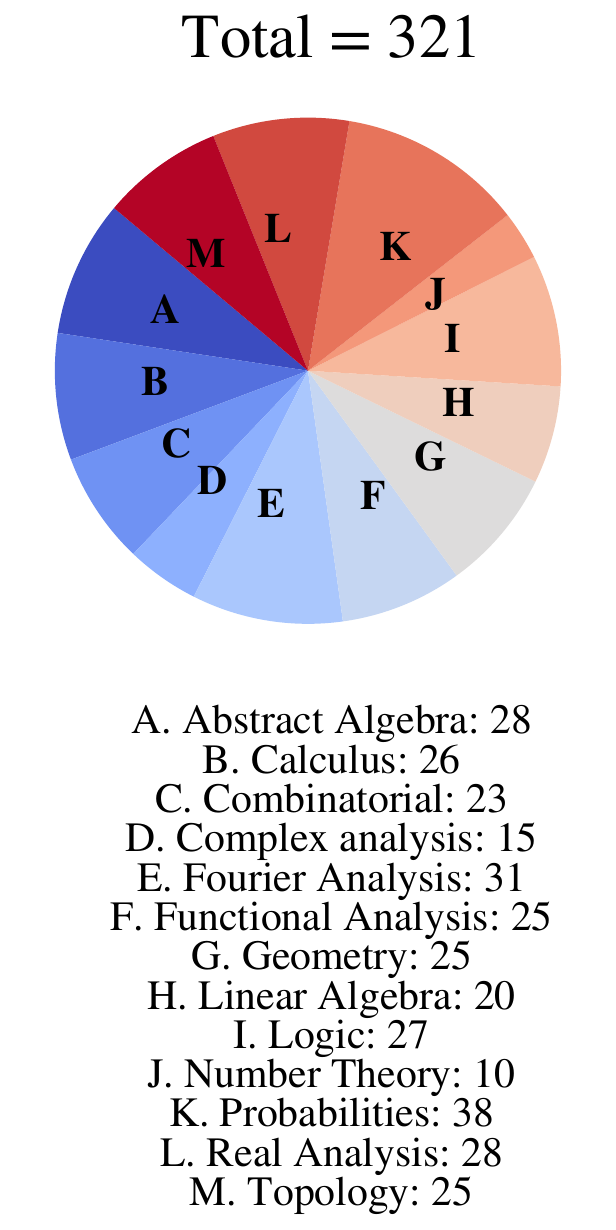}
\end{minipage}
\hfill
\begin{minipage}[t]{0.65\linewidth}
\vspace{0pt}
\centering
\setlength{\fboxsep}{4pt}
\fcolorbox{black}{gray!8}{%
\begin{minipage}{0.93\linewidth}
\fontsize{7.5pt}{8.5pt}\selectfont
\setlength{\abovedisplayskip}{0.25em}
\setlength{\belowdisplayskip}{0.25em}
\refstepcounter{example}
\label{ex:example_probability_diagnostics}
\textbf{Example~\theexample: Diagnostic tasks for a probability problem.}

\smallskip
\noindent\textbf{Original problem.}
Let $S_n=X_1+\dots+X_n$ be a random walk, where the $X_i$ are independent and identically distributed with
$\mathbb{P}(X_i=1)=p$, $\mathbb{P}(X_i=-1)=1-p$, and $p\neq\frac{1}{2}$.
For integers $a\leq-1$ and $b\geq1$, let
\[
\tau=\min\{n\geq1:S_n=a\text{ or }S_n=b\}.
\]
Show that
\[
\mathbb{E}\tau=
\frac{a(1-r^b)+b(r^a-1)}{(r^a-r^b)(2p-1)},
\qquad r=\frac{1-p}{p}.
\]

\smallskip
\noindent\textbf{Direct-answer problem.}
Under the same assumptions, compute $\mathbb{E}\tau$.

\smallskip
\noindent\textbf{Multiple-choice reasoning question.}
Which result or concept is most useful for solving this problem?
\begin{enumerate}[label=(\alph*),leftmargin=2em,itemsep=0pt,topsep=0.2em]
    \item Doob decomposition
    \item Doob's martingale convergence theorems
    \item Optional stopping theorem
    \item Limiting distribution of a Markov chain
    \item Distribution of arrival times for Poisson processes
\end{enumerate}
\end{minipage}}
\end{minipage}
\vspace{-0.4em}
\caption{Left: Distribution of the 321 problems across mathematical domains. Right: An original probability problem with its direct-answer and multiple-choice diagnostic tasks.}
\label{fig:dist_by_category_n_example}
\vspace{-0.75em}
\end{figure}

\benchname\ contains 321 problems drawn from undergraduate- and graduate-level mathematics textbooks or contributed by domain experts. The problems span 13 domains---number theory, linear algebra, abstract algebra, calculus, real analysis, complex analysis, Fourier analysis, functional analysis, probability, topology, geometry, combinatorics, and logic---and are selected to maximize coverage while minimizing redundancy. We formalize 298 problems in Lean 4; the remaining 23 are retained for auxiliary evaluations and deferred for future formalization because of current gaps in Mathlib. For each suitable problem, we construct up to three auxiliary tasks: a direct-answer problem, a multiple-choice reasoning question, and an expert-crafted transformed problem. Figure~\ref{fig:dist_by_category_n_example} shows the domain distribution alongside an example of the diagnostic tasks.

% \begin{figure}[h]
%     \centering
%     \includegraphics[width=\linewidth]{Figures/example.jpeg}
%     \caption{Example of \benchname~with three in-depth corresponding questions accompanying the natural language statement. {\color{blue} xiaoyu can you take a look how to make them more presentable}}
%     \label{fig:example}
% \end{figure}

\textbf{Direct-answer problems.}
We recast suitable proof problems as direct-answer questions that ask the model to compute a target quantity or expression without writing a Lean proof. These questions test whether the model can solve the underlying mathematics in natural language, thereby helping separate informal problem solving from formal proof construction. Their domain distribution is shown in Figure~\ref{fig:computational_by_category}.

\textbf{Multiple-choice problems.}
We ask models to identify the theorem, concept, or reasoning strategy most relevant to the original problem from several plausible options. These questions probe whether a model has the background knowledge needed to approach the proof in natural language. We construct MC questions for 293 problems, excluding those whose solutions rely primarily on direct computation or basic facts rather than a central theorem or proof strategy. Their domain distribution is shown in Figure~\ref{fig:mc_by_category}.
An example of direct-answer and multiple-choice questions is shown in Example~\ref{ex:example_probability_diagnostics}.
% \xl{for example, there we add the same format of the figure as 3.1, saying that this is for the algorithm to pick the correct theorem / underlying method., same for direct answer and natural language.}

\textbf{Transformed problems.}
Domain experts manually reformulate selected problems so that they appear substantially different while preserving the same underlying mathematical reasoning. Comparing performance across the original and transformed versions tests whether models reason consistently rather than relying on familiar wording or memorized patterns~\cite{mirzadeh2025gsmsymbolic}.
% \xl{This is a perfect example and it is very clear that the transformed question is: different questions relying on the same mathematical ideas. I suggest we remove the previous 4 subfigures and add something like this to the other sections. }

Example~\ref{ex:example_real_analysis} illustrates such a transformation. Although the two formulations appear distinct, they rely on the same mathematical ideas, and recognizing this connection requires a deep understanding of the underlying concepts. In total, we construct transformed versions of 30 problems, with their distribution across domains shown in Figure~\ref{fig:transformed_by_category}.

\begin{center}
\setlength{\fboxsep}{6pt}
\fcolorbox{black}{gray!8}{%
\begin{minipage}{0.94\linewidth}
\footnotesize
\linespread{0.9}\selectfont
\setlength{\abovedisplayskip}{0.35em}
\setlength{\belowdisplayskip}{0.35em}
\setlength{\abovedisplayshortskip}{0.2em}
\setlength{\belowdisplayshortskip}{0.2em}
\refstepcounter{example}
\label{ex:example_real_analysis}
\textbf{Example~\theexample: Transformed questions in real analysis.}

\smallskip
\noindent\textbf{Natural language problem.}
{\scriptsize
Let $S = \{(0,0), (2,0), (0,1)\}$. Prove that
\[
\{\lambda_1(0,0) + \lambda_2(2,0) + \lambda_3(0,1) :
\lambda_1, \lambda_2, \lambda_3 \geq 0,
\lambda_1 + \lambda_2 + \lambda_3 = 1\}
=
\{(x,y) \in \mathbb{R}^2 : x \geq 0, y \geq 0, \frac{x}{2} + y \leq 1\}.
\]}

\noindent\textbf{Transformed problem.}
{\scriptsize
Let $S = \{(0,0), (2,0), (0,1)\}$. Prove that the smallest convex set containing $S$ is
\[
\{(x,y) \in \mathbb{R}^2 : x \geq 0, y \geq 0, x/2 + y \leq 1\}.
\]}
\end{minipage}}
\end{center}

% \subsection{Data collection}
% \subsubsection*{Data collection} 
% \xl{the first part of this should be merged into previous section.}
% The  are drawn from undergraduate- and graduate-level mathematics textbooks or contributed by domain experts. 
% Problems in \benchname\ are selected to provide broad coverage while minimizing redundancy and to expose meaningful failure modes in contemporary LLMs. When candidate problems involve similar concepts, we prioritize those that emphasize distinct aspects of the underlying theory. Contributors also interacted with models such as ChatGPT, Gemini, and DeepSeek to assess the diagnostic value of candidate problems. A complete list of sources is provided in Appendix~\ref{app:data_sources}.
%
% All contributors are PhD students who have completed graduate coursework or conduct research in the relevant field. Each problem statement and its auxiliary questions were independently reviewed by a second expert. We do not label problems as undergraduate- or graduate-level because this distinction is subjective and varies across institutions and curricula.

\textbf{Data collection.}
To provide broad coverage while minimizing redundancy and exposing meaningful model failure modes, we prioritize problems that emphasize distinct aspects of the underlying theory when several candidates involve similar concepts. Contributors also interact with models such as ChatGPT, Gemini, and DeepSeek to assess each problem's diagnostic value. All contributors are PhD students who have completed graduate coursework or conduct research in the relevant domain, and a second expert independently reviews every problem statement and its auxiliary questions. Because the distinction between undergraduate- and graduate-level problems is subjective and varies across institutions and curricula, we do not assign these labels. A complete list of sources is provided in Appendix~\ref{app:data_sources}.

\subsection{Human-in-the-loop autoformalization and challenges}
\label{sec:autoformalization_pipeline}
\begin{figure}[h!]
    \centering
    \includegraphics[width=\linewidth]{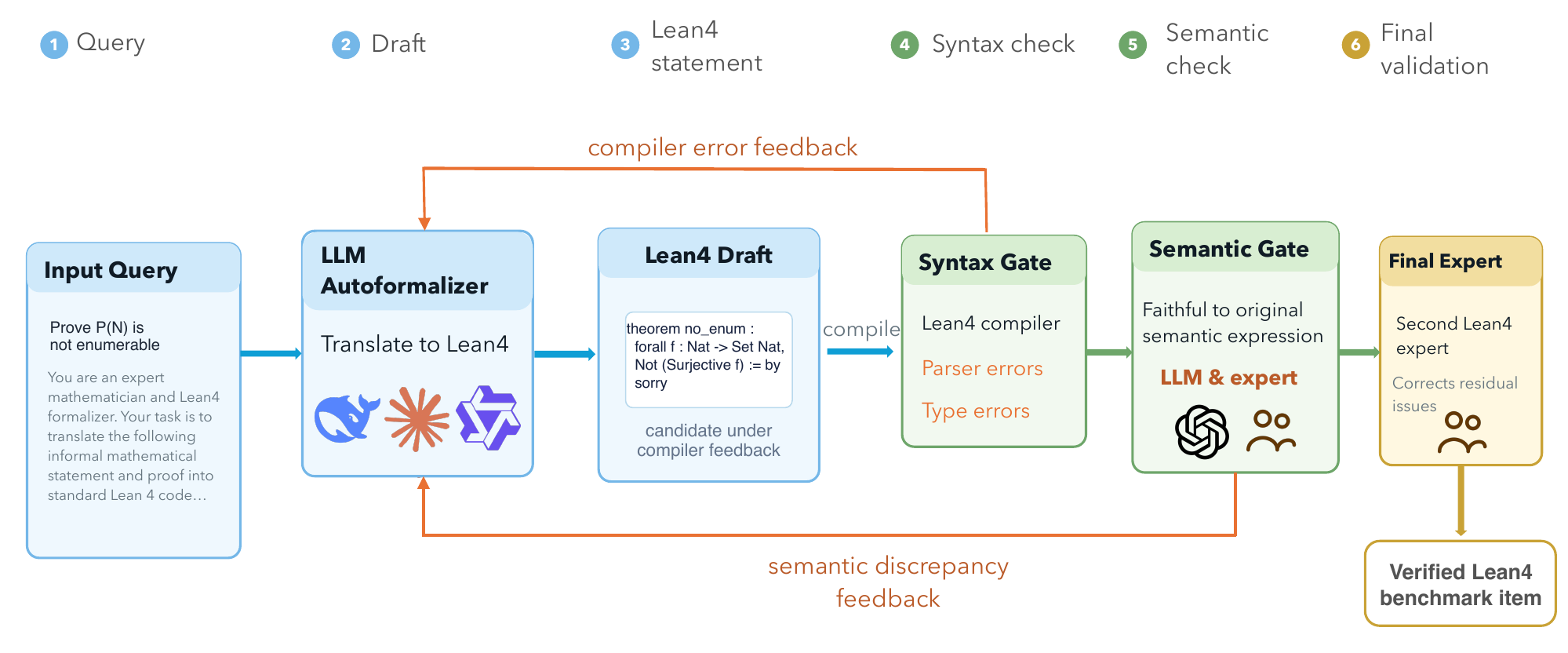}
    \vspace{-2em}
    \caption{Human-in-the-loop autoformalization process for \benchname.}
    \label{fig:formalization}
    \vspace{-0.8em}
\end{figure}

To formalize the natural-language problems in \benchname, we use the LLM-assisted, human-in-the-loop pipeline illustrated in Figure~\ref{fig:formalization}. This design compensates for the inability of current LLMs to independently ensure both valid Lean 4 code and semantic fidelity to the original problems.

Following an interactive, feedback-intensive workflow guided by human experts, the process comprises four stages: \textit{Initial Autoformalization, Syntax Correction, Dual Semantic Verification}, and \textit{Final Expert Review}. An LLM first generates a Lean 4 statement, and compiler errors are returned to the model for iterative correction until the statement type-checks. The resulting formalization is then checked for semantic fidelity by an independent LLM and a human expert, followed by a final review by a second expert. Further details are provided in Appendix~\ref{app:autoformalization}. 
This design choice is motivated by two main considerations: the current strengths and limitations of LLMs in mathematical formalization and the distinctive complexity of \benchname.

% Our formalization process involves a LLMs assisted human-in-the-loop pipeline. 
% For each problem, we first employ LLM-based autoformalizers to translate the problem statements from natural language statement into Lean 4. The prompt for this step of formalization is shown in Appendix~\ref{app:autoformalization}. 
% Despite the advancement of the LLMs, formalization solely with LLMs is far from reliable at current stage. To correct syntax and semantic errors, we implemented a two-step error correction procedures.
% The syntax errors are first checked by the Lean 4 compiler and the messages from the complier is provided to the autoformalizers for improvement until no syntax error is detected.
% To ensure that the Lean 4 theorem statement represent the statement faithfully, we conduct semantic checking by both human experts and an additional LLM-based autoformalizer. 
% If any discrepancy is detected, the feedback will be provided to the autoformalizer to improve on the Lean 4 theorem statement. 
% Once the Lean 4 statement is error-proof, a second human expert conducts a final check and correction.
% The process of autoformalization with the aid of LLMs is illustrated in Figure~\ref{fig:formalization}.

% Because \benchname~covers a wide range of topics as well as difficulty levels, not every question can be formalized in Lean 4 with the current status of the package. 

% Hence our formalization process involves more expert’s labor. We formalize the questions with the goal that they explicitly express the original statement while complying with the syntax in Lean 4. For transformed problems with only rephrasing, we tried our best to preserve the concepts in the original natural language statements. 

% While fully automated translation pipelines have shown great promise in contemporary autoformalization benchmarks, we opted for a more interactive and feedback-intensive approach.
% % human-intensive approach. 
% This design choice is guided by two main considerations: the subtle nature of mathematical formalization and the distinct complexity of \benchname.

First, we observe that general-purpose LLMs can often explain the mathematical meaning of the code and identify semantic issues, but still incorrectly predict that the code does not compile. However, without actually running the Lean compiler, they are much less reliable at judging whether the code will be accepted by the Lean type checker.
Using James Hanson’s Lean 4 ``junk theorems'', semantically misleading snippets that nevertheless compile, we test whether models can predict compilation and explain the code's meaning~\cite{hanson_junk_theorems_in_lean}. The results show that LLMs often misjudge what Lean accepts; full details are provided in Appendix~\ref{app:junk-theorems}.

% When given Lean 4 code that compiles, these models can often explain the mathematical meaning of the code and identify semantic issues. However, without actually running the Lean compiler, they are much less reliable at judging whether the code will be accepted by the Lean type checker.

% We illustrate this gap using James Hanson’s collection of Lean 4 “junk theorems”~\cite{hanson_junk_theorems_in_lean} as a diagnostic testbed. These examples are especially useful because they look suspicious or mathematically misleading, but nevertheless compile in Lean. We test  Each model is asked two questions for every self-contained Lean snippet: whether the snippet compiles, and what theorem it proves. This setup allows us to distinguish semantic understanding from type-checking judgment. Full details and results are provided in Appendix~\ref{app:junk-theorems}.

% As shown in Table~\ref{tab:junk_results_main}, the models can often explain the formal trick behind a junk theorem, but still incorrectly predict that the code does not compile. This suggests that LLMs do not reliably know where the boundary lies between Lean code that merely appears plausible and Lean code that is actually accepted by the type checker.

This observation motivates the design of our formalization pipeline. We use LLMs for the tasks where they are relatively strong, such as checking whether a formal statement matches the intended mathematical meaning. At the same time, we do not rely on them as syntax or type-checking judges. Instead, we place the Lean verifier in the loop: Lean provides direct compiler feedback on whether the code type-checks, while the LLMs help interpret and revise the formalization at the semantic level.

% First, translating natural language mathematics into interactive theorem provers is a highly nuanced task. 
% Even advanced LLMs have a poor sense of what the Lean type checker actually accept without being allowed to run code directly.
% We use James Hanson’s collection of Lean 4 ``junk theorems”~\cite{hanson_junk_theorems_in_lean} as a diagnostic testbed for evaluating whether LLMs can distinguish between a statement’s misleading informal reading and its actual formal validity. 
% Each model is asked both whether a self-contained Lean snippet compiles and what theorem it proves, allowing us to measure the gap between semantic understanding and type-checking judgment. {\color{blue} See Appendix xxx}.
% We observe that models can often explain the formal trick behind a junk theorem, but still wrongly predict that the Lean code does not compile. 
% This suggests that LLMs often misjudge the boundary between Lean code that merely looks plausible and Lean code that actually type-checks.
% Therefore, in our formalization pipeline, we use LLMs within a verifier-in-the-loop framework: Lean feedback compensates for their weak calibration to the type checker, while the models are still leveraged for semantic checking, where they have shown stronger ability to identify whether a formal statement preserves the intended mathematical meaning.

% the advanced difficulty and diverse domain coverage of \benchname\ mean that many problems cannot be easily automated. 
A second major obstacle to fully automated translation of \benchname\ is the limited availability of prerequisite definitions and lemmas in Mathlib at the time of writing. Because \benchname\ includes problems from advanced and specialized areas of mathematics, some of the basic objects needed to state these problems have not yet been formalized in Lean by the open-source community.
For example, some candidate problems in Riemannian geometry cannot yet be stated or proved in Lean because Mathlib lacks the required specialized definitions and supporting results, such as the Bishop--Gromov inequality. This illustrates how gaps in the current formal library constrain immediate formalization; Appendix~\ref{app:mathlib-coverage} provides a detailed example.

% This example illustrates why full automation is currently unrealistic for parts of \benchname. The difficulty is not only that the statements are mathematically advanced, but also that the formal infrastructure required to express and prove them is missing. As a result, our pipeline requires human expert involvement to identify unavailable prerequisites, decide how to represent advanced concepts, and determine when a problem is outside the current scope of Mathlib.

% This problem serves as specific instance of a general weakness of matlib in various fields. While Matlib has robust foundations in various fields such as  combinatorics, number theory and algebraic topology, we find its advanced geometric foundations to be comparably under developed. 

To assess whether problems were immediately formalizable, we systematically surveyed Mathlib using its official documentation and LeanSearch, supplemented by Loggle when the required types were known. Expert judgment remains necessary to identify structural gaps and determine whether a translation is viable. Problems requiring prohibitive upstream library development are retained for auxiliary evaluations and deferred to future iterations of \benchname\ as the Lean 4 ecosystem matures.

% {\color{red} After our challenging, mention why formalization is challenging by itself. Our process replies heavily on human discretion even with the aid of autoformalizers. }

% For problems that cannot be formalized in Lean 4 yet, we do not discard them because we not only examine models’ capability through formal proof questions but a variety of venues. Moreover, though they cannot be formalized yet in Lean 4 does not mean that with the evolvement of the package and the development, they can be easily updated into high quality benchmark problems. 

% {\color{red} TO DO: 1. polish autoformalization and challenges, we don't want to emphasize this part but mention how hard it is and why we didn't use more automated pipeline for this task. 2. setting up a experiment to test on whether LLM identifies them and their reaction.
% 3. Email Connor once section 4 \& 5 are done. }

\section{Diagnostic evaluation of theorem provers}
We evaluate a diverse set of theorem-proving and general-purpose LLMs on \benchname. 
Evaluated models and protocols are described in Section~\ref{sec:setting}.
Section~\ref{sec:theorem_provers_results} measures end-to-end Lean 4 proof construction. Section~\ref{sec:DA_MC} assesses natural-language problem solving and knowledge of relevant theorems and proof strategies. Section~\ref{sec:transformations_results} evaluates whether models remain robust under mathematically equivalent reformulations. Together, these evaluations provide a fine-grained view of model capabilities, revealing whether failures arise from mathematical reasoning, background knowledge, formal proof construction, or sensitivity to problem formulation.

\subsection{Settings}
\label{sec:setting}
We evaluate two types of proof-generation approaches on \benchname: proof-step generation models and whole-proof generation models.
Proof-step generation based models usually utilize verifier to obtain information on the current tactic state and often times construct valid proofs with tree-searching schemes.  
For this category, we evaluate \benchname\ on InternLM2-Math-Plus-7B \cite{ying2024internlmmath}, InternLM2-Step-Prover\cite{wu2024internlm25stepproveradvancingautomatedtheorem}, DeepSeek-Prover-V1.5-RL + RMaxTS \cite{xin2024deepseekproverv15harnessingproofassistant}, DeepSeek-Prover + MA-LoT, Goedel-Prover + MA-LoT \cite{wang2025malot}, Goedel-Prover-V2-32B \cite{lin2025goedelproverv2scalingformaltheorem}.

Whole-proof generation models in contrast produce an entire proof code given the prompts, without involving communication between prover model and the verifier. The evaluation metrics used for whole-proof generation models are usually pass@K by measuring the number of problems for which at least one valid proof is found among the top K generated attempts.
From this category, we evaluated on Goedel-Prover-SFT, Goedel-Prover-DPO \cite{lin2025goedelproverfrontiermodelopensource},
DeepSeek-Prover-V1.5, DeepSeek-Prover-V1.5-SFT and DeepSeek-Prover-V1.5-RL\cite{xin2024deepseekproverv15harnessingproofassistant}.

In addition to the open-sourced models, we also evaluate \benchname~on GPT-5.4 \cite{singh2025openaigpt5card} Claude-sonnet-5~\cite{anthropic2026claudesonnet5}, DeepSeek-V3.2\cite{deepseekai2025deepseekv32pushingfrontieropen}, DeepSeek-R1 \cite{Guo_2025}.
We further evaluate three AI agents with iterative theorem-proving capabilities: Opengauss~\cite{mathinc2026opengauss} 
% Aristotle~\cite{harmonic2025aristotle}, 
and Codex with GPT-5.4. These agents leverage feedback from the Lean environment to iteratively generate and refine formal proofs.
For details on the computational budget used for evaluation of each model, see Appendix~\ref{sec:app_comp_budget}.

\subsection{Formal theorem-proving performance}
\label{sec:theorem_provers_results}
\begin{figure}[h]
    \centering
    \begin{minipage}[t]{0.46\textwidth}
        \vspace{20pt}
        \centering
        \resizebox{\linewidth}{!}{\scriptsize
\begin{tabular}{lc}
\hline
Model & Accuracy (\%) \\
\hline
DeepSeek-Prover-V1.5 Base & 1.25 \\
DeepSeek-Prover-V1.5 SFT & 9.06 \\
DeepSeek-Prover-V1.5 RL & 11.25 \\
DeepSeek-Prover-V1.5 RL + RMaxTS & 16.56 \\
InternLM2-Math-Plus-7B & 13.12 \\
InternLM2.5-StepProver & 10.31 \\
Goedel-Prover-SFT & 10.62 \\
Goedel-Prover-DPO & 11.25 \\
Goedel-Prover-V2 & 21.88 \\
DeepSeek-Prover + MA-LoT & 10.00 \\
Goedel-Prover + MA-LoT & 11.88 \\
GPT-5.4 & 10.62 \\
DeepSeek-R1 & 10.94 \\
DeepSeek-V3.2 & 5.31 \\
\hline
\end{tabular}}
        \vspace{0.5em}
        \captionof{table}{\footnotesize Performance comparison of theorem prover LLMs on \benchname~ in Lean 4.}
        \label{tab:all-models-lean4-accuracy}
    \end{minipage}
    \hfill
    \begin{minipage}[t]{0.5\textwidth}
        \vspace{0pt}
        \centering
        \includegraphics[width=\linewidth]{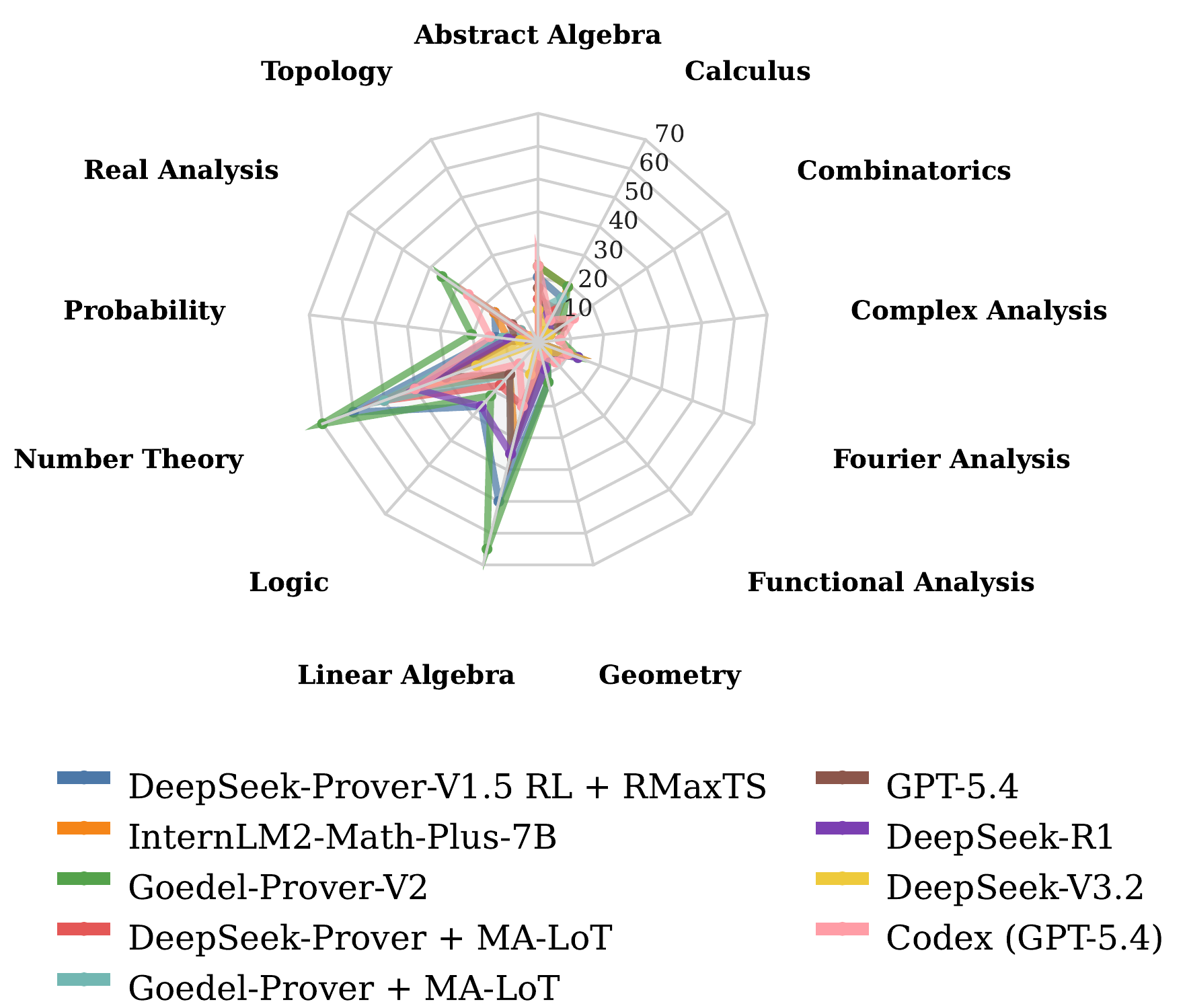}
        \captionof{figure}{\footnotesize Accuracy by domain in \benchname\ for selected models.}
        \label{fig:lean4_domain}
    \end{minipage}
    \vspace{-0.5em}
\end{figure}

Table~\ref{tab:all-models-lean4-accuracy} presents the performance of theorem-proving LLMs on \benchname\ in Lean 4. Overall, theorem-proving-specific training and search improve Lean 4 proving performance, but the overall success rate remains limited on \benchname.

Among DeepSeek-Prover-V1.5 variants, accuracy increases from 1.25\% for the base model to 11.25\% with reinforcement learning, and further to 16.56\% with RMaxTS, suggesting the benefit of search-based inference in the sparse-reward setting of formal theorem proving. We also observe smaller gains from the MA-LoT framework across both DeepSeek-Prover and Goedel-Prover. In contrast, general-purpose LLMs such as GPT-5.4 and DeepSeek-R1 achieve only modest accuracy despite their larger scale and broad mathematical knowledge, highlighting the gap between mathematical reasoning and producing successful proofs in Lean 4.

Goedel-Prover-V2 achieves the best performance at 21.88\%, followed by DeepSeek-Prover-V1.5 RL + RMaxTS. Notably, both systems employ verifier-guided search, treating theorem proving as an interactive process with Lean rather than a one-shot generation task. Their strong performance highlights the importance of effective interaction with the proof assistant for end-to-end theorem proving.

\textbf{Unbalanced performance across different domains.}

We also observe substantial variation across mathematical domains, as shown in Figure~\ref{fig:lean4_domain}. Goedel-Prover-V2 remains strongest in number theory and linear algebra. Codex (GPT-5.4) shows a different profile: it performs particularly well in number theory, real analysis, and abstract algebra, and obtains the highest accuracy among the evaluated systems in combinatorics and functional analysis. Nevertheless, all evaluated models, including Codex, score 0\% on topology.
These domain-level differences may reflect both the composition of model training data and the availability of relevant formal library resources. The concentration of Goedel-Prover-V2 and DeepSeek-Prover variants in number theory and linear algebra may be related to their emphasis on formal and competition-style mathematics during training~\cite{lin2025goedelproverfrontiermodelopensource,xin2024deepseekproverv15harnessingproofassistant}. Codex improves the overall accuracy of GPT-5.4, but the improvement is not uniform across domains. This mixed result suggests that agentic execution may be able to help overcome some failures in library use and proof construction, but does not consistently remove domain-specific weaknesses. The performance of each model across all domains can be found in Tables~\ref{tab:deepseek-v15-base}-\ref{tab:codex-gpt5.4-domain} in Appendix~\ref{app:additiona_results}.

\subsection{Direct-answer and multiple-choice mathematical reasoning performance}
\label{sec:DA_MC}

We evaluate models on direct-answer (DA) and multiple-choice (MC) problems to isolate natural-language mathematical reasoning and background knowledge in our assessment. Table~\ref{tab:computational_mc_accuracy} reports the results. We exclude systems such as MA-LoT that do not support DA or MC answer generation. Details of DA and MC answer generation and evaluation are provided in Appendices~\ref{app:evaluation_computational} and~\ref{app:evaluation_MC}.

\textbf{Direct-answer problems.}
We observe that most models achieve higher accuracy on DA problems than on the corresponding Lean 4 theorem-proving tasks. This gap indicates that models can often solve the underlying mathematical problem, but still struggle to express the solution as a formal proof accepted by Lean.
Goedel-Prover is a notable exception: despite its strong theorem-proving performance, it performs comparatively worse on DA tasks, suggesting that its strengths are more closely tied to Lean proof search and theorem application than to general mathematical answer generation in natural language. 
% Overall, the DA benchmark helps separate informal mathematical problem-solving ability from end-to-end formalization ability in Lean 4.

% Table~\ref{tab:computational_accuracy} reports model performance on the direct-answer benchmark. With the exception of Goedel-Prover, all evaluated models achieve substantially higher accuracy than on the corresponding Lean 4 theorem-proving tasks. 
% % This result suggests that mathematical reasoning and computation are considerably easier than formal proof construction. 
% Notably, Goedel-Prover performs comparatively worse on these tasks despite its strong theorem-proving ability, indicating a greater degree of specialization toward interactive proof construction and theorem application within Lean. 
% In contrast, DeepSeek-Prover and InternLM models retain stronger general-purpose mathematical reasoning and symbolic computation capabilities, likely inherited from their large language model foundations. Overall, direct-answer problems appear to favor broad mathematical reasoning abilities over specialization in formal proof search.

% \subsection{Multiple-Choice Mathematical Knowledge and Reasoning Performance}
% \label{sec:MC-results}
\textbf{Multiple-choice problems.}
On MC problems, we also see that most models achieve substantially higher accuracy than Lean 4 theorem-proving accuracy. 
Goedel-Prover remains an exception, consistent with its stronger specialization toward Lean proof search than general answer selection.
% We observe that DeepSeek-R1 underperforms DeepSeek-V3.2 on MC questions primarily because of answer omission, with 141 of R1's 153 errors being empty predictions compared with only three for V3.2; meanwhile, DeepSeek-Prover-V1.5 SFT and RL score below the base model, suggesting that optimization for formal proving may trade off some broader mathematical reasoning behavior.
Together with the DA results, these findings reinforce that formal proof construction involves challenges beyond mathematical background knowledge, natural-language reasoning, and answer generation.
\begin{figure}[h!]
    \centering
    \vspace{-1em}
    \begin{minipage}[t]{0.45\textwidth}
        \vspace{20pt}
        \centering
        \resizebox{\linewidth}{!}{\begingroup
\scriptsize
\renewcommand{\tabularxcolumn}[1]{m{#1}}
\renewcommand{\arraystretch}{0.93}
\begin{tabularx}{\textwidth}{@{}>{\raggedright\arraybackslash}Xrr@{}}
\toprule
Model & DA (\%) & MC (\%) \\
\midrule
DeepSeek-Prover-V1.5 Base & 25.9 & 53.08 \\
DeepSeek-Prover-V1.5 SFT & 32.1 & 32.53 \\
DeepSeek-Prover-V1.5 RL & 37.0 & 33.90 \\
DeepSeek-Prover-V1.5 RL + RMaxTS & 42.0 & 38.70 \\
InternLM2-Math-Plus-7B & 39.5 & 64.04 \\
Goedel-Prover-SFT & 4.9 & 21.23 \\
Goedel-Prover-DPO & 4.9 & 19.52 \\
DeepSeek-R1 & 58.0 & 51.71 \\
DeepSeek-V3.2 & 66.7 & 75.00 \\
GPT-5.4 & 64.2 & 82.88 \\
\bottomrule
\end{tabularx}
\endgroup
}
        \captionof{table}{\small Overall accuracy on direct-answer and multiple-choice reasoning questions.}
        \label{tab:computational_mc_accuracy}
    \end{minipage}
    \hfill
    \begin{minipage}[t]{0.5\textwidth}
        \vspace{0pt}
        \centering
        \resizebox{\linewidth}{!}{% \begin{table}[htbp]
% \centering
% \footnotesize
% \caption{Original vs. transformed comparison on Lean 4 questions across models. Entries are counts only.}
% \label{tab:lean4-transformation-comparison}
% \begin{tabular}{lcc}
% \hline
% Model & M$\checkmark$/T$\times$ & M$\times$/T$\checkmark$ \\
% \hline

% % Open-source models
% DeepSeek V1.5 Base & 0 & 0 \\
% DeepSeek V1.5 SFT & 2 & 2 \\
% DeepSeek V1.5 RL & 4 & 2 \\
% RMaxTS & 4 & 2 \\
% InternLM2-Math & 2 & 2 \\
% InternLM2.5 Step Prover & 3 & 2 \\
% LoT DeepSeek V1.5 & 4 & 2 \\
% LoT Goedel & 4 & 2 \\
% Goedel SFT & 4 & 2 \\
% Goedel DPO & 4 & 2 \\
% Goedel V2 & 6 & 0 \\

% % Non open-source models
% GPT-5.4 pass@16 & 2 & 1 \\
% DeepSeek Reasoner pass@16 & 4 & 0 \\
% DeepSeek API Chat pass@32 & 3 & 1 \\

% \hline
% \end{tabular}
% \end{table}

\scriptsize
\begin{tabular}{lcc}
\hline
Model & O$\checkmark$/T$\times$ & O$\times$/T$\checkmark$ \\
\hline

DeepSeek-Prover-V1.5 Base & 0 & 0 \\
DeepSeek-Prover-V1.5 SFT & 2 & 2 \\
DeepSeek-Prover-V1.5 RL & 4 & 2 \\
DeepSeek-Prover-V1.5 RL + RMaxTS & 4 & 2 \\
InternLM2-Math-Plus-7B & 2 & 2 \\
InternLM2.5-StepProver & 3 & 2 \\
DeepSeek-Prover + MA-LoT & 4 & 2 \\
Goedel-Prover + MA-LoT & 4 & 2 \\
Goedel-Prover-SFT & 4 & 2 \\
Goedel-Prover-DPO & 4 & 2 \\
Goedel-Prover-V2 & 6 & 0 \\

GPT-5.4 & 2 & 1 \\
DeepSeek-R1 & 4 & 0 \\
DeepSeek-V3.2 & 3 & 1 \\

\hline
\end{tabular}

% \scriptsize
% \begin{tabular}{lcccc}
% \hline
% Model & O$\checkmark$/T$\checkmark$ & O$\checkmark$/T$\times$ & O$\times$/T$\checkmark$ & O$\times$/T$\times$ \\
% \hline

% DeepSeek-Prover-V1.5 Base & 0 & 0 & 0 & 22 \\
% DeepSeek-Prover-V1.5 SFT & 1 & 2 & 2 & 17 \\
% DeepSeek-Prover-V1.5 RL & 1 & 4 & 2 & 15 \\
% DeepSeek-Prover-V1.5 RL + RMaxTS & 2 & 4 & 2 & 14 \\
% InternLM2-Math-Plus-7B & 2 & 2 & 2 & 16 \\
% InternLM2.5-StepProver & 1 & 3 & 2 & 16 \\
% DeepSeek-Prover + MA-LoT & 1 & 3 & 2 & 16 \\
% Goedel-Prover + MA-LoT & 1 & 4 & 2 & 15 \\
% Goedel-Prover-SFT & 1 & 4 & 2 & 15 \\
% Goedel-Prover-DPO & 1 & 4 & 2 & 15 \\
% Goedel-Prover-V2 & 2 & 6 & 0 & 14 \\

% GPT-5.4 & 1 & 2 & 1 & 18 \\
% DeepSeek-R1 & 1 & 5 & 0 & 16 \\
% DeepSeek-V3.2 & 1 & 3 & 1 & 17 \\

% \hline
% \end{tabular}
}
        \captionof{table}{\footnotesize Original vs. transformed comparison on theorem proving questions. Entries are counts only.}
        \label{tab:lean4-transformation-comparison}
    \end{minipage}
    \vspace{-1em}
\end{figure}

\subsection{Robustness to mathematically equivalent transformations}
\label{sec:transformations_results}
A distinctive feature of \benchname\ is its transformed theorem-proving problems in both natural language and Lean, which domain experts manually construct to preserve the underlying mathematical reasoning while changing the problem formulation.
Table~\ref{tab:lean4-transformation-comparison} compares model performance on the original and transformed versions. We use O$\checkmark$/T$\times$ to denote cases where the original version is solved but the transformed version is not, and O$\times$/T$\checkmark$ to denote the reverse.

The results reveal a consistent robustness gap. For most models, O$\checkmark$/T$\times$ is larger than O$\times$/T$\checkmark$, meaning that models more often solve the original problem but fail on an equivalent transformed version than vice versa. 
This effect is especially pronounced for Goedel-Prover-V2 and DeepSeek-R1, which show multiple O$\checkmark$/T$\times$ cases but no O$\times$/T$\checkmark$ cases. 
Because the transformations preserve the underlying mathematics, failures on reformulated versions show that current theorem provers depend on how a problem is presented. Changes in wording or structure can disrupt learned proof patterns, theorem retrieval, or proof-search strategies. Although the absolute counts are small, the same asymmetry appears across model families, motivating training that rewards consistency across equivalent formulations rather than reliance on surface cues.
Overall, these evaluations show that current models are limited not only by mathematical knowledge, but also by formal proof construction, uneven domain coverage, and sensitivity to equivalent problem formulations. 
% To identify actionable ways to address these limitations, the following section qualitatively examines failure modes, successful proof patterns, and the effects of verifier feedback and natural language hints.

% Most models exhibit a clear asymmetry between these two counts: O$\checkmark$/T$\times$ is consistently larger than O$\times$/T$\checkmark$. In other words, models are more likely to solve the original problem but fail on a mathematically equivalent transformed version than vice versa. 
% % Since the transformed problems preserve the underlying mathematical reasoning, an ideal model with robust semantic understanding should maintain correctness across both versions. 
% The observed asymmetry therefore suggests that current theorem provers remain sensitive to problem formulation, with transformations disrupting learned proof patterns, theorem retrieval, or proof-search strategies. This effect is particularly pronounced for Goedel-Prover-V2 and DeepSeek-R1, both of which exhibit multiple O$\checkmark$/T$\times$ cases but no O$\times$/T$\checkmark$ cases. Although the absolute counts are small due to the overall difficulty of theorem proving in Lean 4, the consistent asymmetry across model families highlights a notable robustness limitation.

% \subsubsection{Results by asking the equivalence}

\section{Failure analysis and improvement strategies}
\label{sec:qualitative_analysis}

In this section, we compare failed and successful proofs to identify what separates plausible attempts from valid Lean proofs in Section~\ref{sec:error_pattern}. 
% Next, Section~\ref{sec:error_domains} examines DeepSeek-R1's chain-of-thought traces across domains to better understand domain-specific reasoning gaps. 
Section~\ref{sec:agent_repair} examines what agentic interaction can and cannot repair. Finally, Section~\ref{sec:reasoning_hints} examines how natural-language reasoning hints affect formal proof construction across model types.

% In this section, we present our in-depth analysis on the performance of existing theorem prover models tested.
% In \ref{sec:error_pattern}, we identify the most salient error from Lean for diagnosis and classification. 
% Section~\ref{sec:success_pattern} analyze on the other hand the success pattern, conducted by human experts.
% Section~\ref{sec:error_domains} analyze the chain-of-the-thoughts from DeepSeek-R1 model, aiming to provide insights for bridging the gap between reasoning in natural language and delivering correct Lean code.
% % We observe that DeepSeek-R1 in many cases is able to identify the correct reasoning but failed to produce a correct Lean code.
% Section~\ref{sec:multi-turn} analyzed how an interactive mode with general-purpose LLMs e.g., DeepSeek-V3.2, DeepSeek-R1 and GPT5.4 change the error pattern. 

\subsection{Success requires precise lean execution}
\label{sec:error_pattern}
\label{sec:success_pattern}

Comparing failed and successful attempts reveals a clear divide. Many failed proofs begin with a reasonable mathematical idea and even resemble valid Lean code, but break because a necessary step is missing, an unsuitable tactic or nonexistent lemma is used, or the proof repeats without progress. Successful proofs avoid these failures by either retrieving the correct Mathlib result or building the argument explicitly, step by step, while tracking the remaining goals. This suggests that improving formal theorem provers requires not only strong mathematical reasoning, but also better control of the proof state, more reliable library retrieval, and the ability to recognize whether each step makes progress. We describe these patterns in more detail below.

\textbf{Error pattern analysis.}
We identify four recurring ways in which proofs fail. First, an \emph{incomplete proof} may follow a reasonable approach but prove only a weaker statement, skip a required case, or treat a difficult step as already finished; Lean therefore still has an unresolved claim at the end (see Appendix~\ref{app:error_example_incorrect_theorem}). Second, a model may choose a proof command that is valid Lean code but cannot establish the current claim. For example, it may use a command intended for linear arithmetic when the proof actually requires reasoning about divisibility (Appendix~\ref{app:error_example_incorrect_tactic}). Third, models sometimes invent names for results that sound as though they belong to Lean's mathematical library but do not actually exist, preventing the proof from being checked. Finally, some proofs repeat the same rewriting or command many times without changing what remains to be proved (Appendix~\ref{app:error_example_repeated}). These failures show that a plausible overall idea is insufficient unless the model also checks what each step proves and whether it moves the proof forward.

\textbf{Successful proof patterns.}
We observe three common ways in which proofs succeed. First, some successful proofs are short because they find an existing result in Lean's mathematical library that closely matches the statement and already captures most of the required argument (Appendix~\ref{app:success_example_lemma}). Second, some proofs translate a step-by-step mathematical argument into Lean by introducing intermediate claims, separating cases, and providing the required objects. Examples \textit{Q4} and \textit{Q5} illustrate this pattern in Appendix~\ref{app:success_example_have}. Third, some problems that appear difficult in ordinary mathematical language become simple after their formal definitions are unfolded; the proof then needs only a suitable example and a few basic checks (Appendix~\ref{app:success_example_trivial}). In contrast to failed attempts, successful proofs either find a result that completes the argument or ensure that each explicit step brings the proof closer to completion.

\subsection{Model-dependent effects of natural-language hints}
\label{sec:reasoning_hints}
We conduct an ablation in which models receive natural-language reasoning hints during Lean 4 proof construction. This experiment is motivated by the results in Section~\ref{sec:DA_MC}: most models can identify relevant mathematical knowledge and proof strategies in multiple-choice questions, yet still struggle with end-to-end theorem proving. We therefore test whether providing this knowledge as a hint can help bridge the gap.
We evaluate DeepSeek-V3.2, DeepSeek-R1, Goedel-Prover-SFT, and Goedel-Prover-DPO with hints derived from the corresponding multiple-choice questions. We use the same theorem-proving setup, adding a standardized hint based on each question and its correct answer to the prompt.

\begin{figure}[h!]
\vspace{-1em}
    \centering
    \begin{minipage}[t]{0.47\textwidth}
        \vspace{0pt}
        \small
        Hints reveal a clear split between general-purpose and proof-specialized models. They improve the two DeepSeek models but substantially reduce the performance of Goedel-Prover-SFT and Goedel-Prover-DPO. This suggests that natural-language reasoning guidance can help general-purpose models, but can interfere with proof-specialized models whose strengths depend more on learned formal proof patterns and Lean-specific theorem application. Prompting and training strategies should therefore be model-dependent: informal reasoning hints may help general-purpose models, whereas proof-specialized systems may benefit more from formal, Lean-aligned guidance.
    \end{minipage}
    \hfill
    \begin{minipage}[t]{0.5\textwidth}
        \vspace{-1pt}
        \centering
        \includegraphics[width=\linewidth,trim=2pt 14pt 0 0,clip]{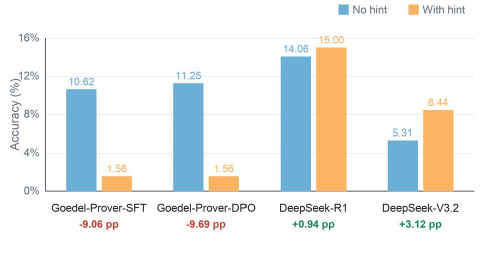}
        \captionof{figure}{\small Performance on Lean 4 formal proving with and without hints from MC questions.}
        \label{fig:lean4-hint-comparison}
        
    \end{minipage}
    \vspace{-0.8em}
\end{figure}

\subsection{What agentic interaction can and cannot repair}
\label{sec:agent_repair}
Building on our error analysis and the benefits of interaction observed in Section~\ref{sec:theorem_provers_results}, we therefore first conduct a controlled experiment that isolates the effect of verifier feedback. Lean error messages are returned to the model for up to $T=3$ turns under a fixed budget $K=N\times T$. Table~\ref{tab:compare_interactive} in the appendix shows that feedback improves both evaluated models. Qualitatively, it is most effective at repairing local syntax and elaboration errors. Once these errors are removed, however, failures often shift to unresolved goals, incorrect theorem applications, and type mismatches. Verifier feedback can therefore correct some local implementation mistakes, but does not reliably repair an incomplete or incorrect proof strategy.

Codex and Gauss extend this interaction with library search and file inspection. Rather than revising the original proof directly, an agent may avoid its failure by finding an applicable Mathlib result or constructing a different formalization. Codex finds valid proofs for 21 problems that non-agentic GPT-5.4 fails to solve, while OpenGauss contributes four additional solves beyond the same baseline. Most successful agent proofs are short and direct that introduce no intermediate lemmas. Only seven Codex proofs and four OpenGauss proofs build chains of three or more intermediate results. This suggests that the agents are most effective when the main mathematical idea is already accessible but must be translated into valid Lean, for example by finding an applicable library theorem or choosing a simpler formalization. They are less successful when a problem requires a new multi-step argument.
\section{Conclusion and Limitations}

In this work, we introduced \benchname, a comprehensive diagnostic benchmark for formal mathematical reasoning in Lean 4 that spans 13 mathematical domains. By moving beyond aggregate theorem-proving accuracy, it provides a fine-grained view of why models succeed or fail. In addition to formal theorem-proving tasks, \benchname includes three auxiliary tasks that allow us to distinguish models' abilities in mathematical problem solving, background knowledge, formal proof construction, and robustness to equivalent reformulations.
% Our evaluation shows that models often identify relevant mathematical ideas but struggle to turn them into complete Lean proofs, highlighting the need for more reliable formal proof construction. Our ablation studies on verifier feedback and natural-language hints further identify promising, model-dependent strategies for training and inference.

% \section{Limitations}

\benchname\ has two main limitations. First, it remains modest in size because creating and validating high-quality problems across many mathematical domains requires substantial expert effort. Second, the benchmark is constrained by the current coverage of Mathlib, so some advanced problems cannot yet be formalized in Lean 4. We plan to formalize these deferred problems as Mathlib expands.

\bibliographystyle{plainnat}
\bibliography{Reference}

\clearpage
\appendix

\section{Related Work}
\label{app:related_work}
\subsection{Existing Formal Mathematics Benchmarks}
\label{app:benchmark_review}
miniF2F~\cite{zheng2021minif2f} evaluates formal proof generation on competition-level mathematics problems, while ProofNet~\cite{azerbayev2023proofnet} provides paired informal and formal statements to study translation from natural language into a proof assistant. FIMO~\cite{liu2023fimo} and PutnamBench~\cite{tsoukalas2024putnambenchevaluatingneuraltheoremprovers} are also derived from mathematical competitions and concentrate largely on algebra, number theory, and analysis. FIMO uses IMO Shortlisted Problems from 2006--2021, whereas PutnamBench draws from the William Lowell Putnam Mathematical Competition.

FormalMath~\cite{yu2025formalmath} expands the scale and domain coverage of formally verified theorems. FormalNumina~\cite{liu2026numina} instead emphasizes structured symbolic and numerical reasoning, focusing on precise formal manipulation. IndiMathBench~\cite{biyani2025indimathbench} extends evaluation to geometry problems involving diagrams and therefore requires both spatial and logical reasoning. Together, these benchmarks provide valuable tests of individual aspects of formal mathematics, but do not jointly diagnose mathematical knowledge, natural-language reasoning, formal proof construction, and robustness to equivalent reformulations.

\subsection{Models for Formal Mathematics}
\label{app:model_review}
One class of methods generates a complete solution or proof in a single pass. GPT-f~\cite{polu2020generative} was an early application of language models to formal proof generation, while Llemma~\cite{azerbayev2023llemma} and InternLM~\cite{ying2024internlmmath} build on stronger pretrained models and mathematical training data. These methods are simple and scalable, but cannot easily recover when an intermediate proof step is incorrect.

Search-based systems instead explore multiple proof paths and use intermediate feedback to revise their choices. Lean-STAR~\cite{lin2024lean}, Real-Prover~\cite{shen2025real}, and Goedel-Prover~\cite{lin2025goedelproverfrontiermodelopensource} employ tree search, best-first search, or structured sampling to find valid proofs. This improves their ability to recover from errors, although it requires greater computation.

Other systems combine large formal training corpora, proof-assistant feedback, or collaborative reasoning. DeepSeek-Prover~\cite{xin2024deepseekproverv15harnessingproofassistant} and TheoremLlama~\cite{wang2024theoremllama} train on extensive formal data; Herald~\cite{gao2024herald} treats proof construction as a sequence of decisions over changing proof states; and MA-LoT~\cite{wang2025malot} uses multiple agents or iterative decomposition to address complex problems.

\section{Data statistics}
\begin{figure}[h!]
    \centering
    \vspace{-1em}\includegraphics[width=0.65\linewidth]{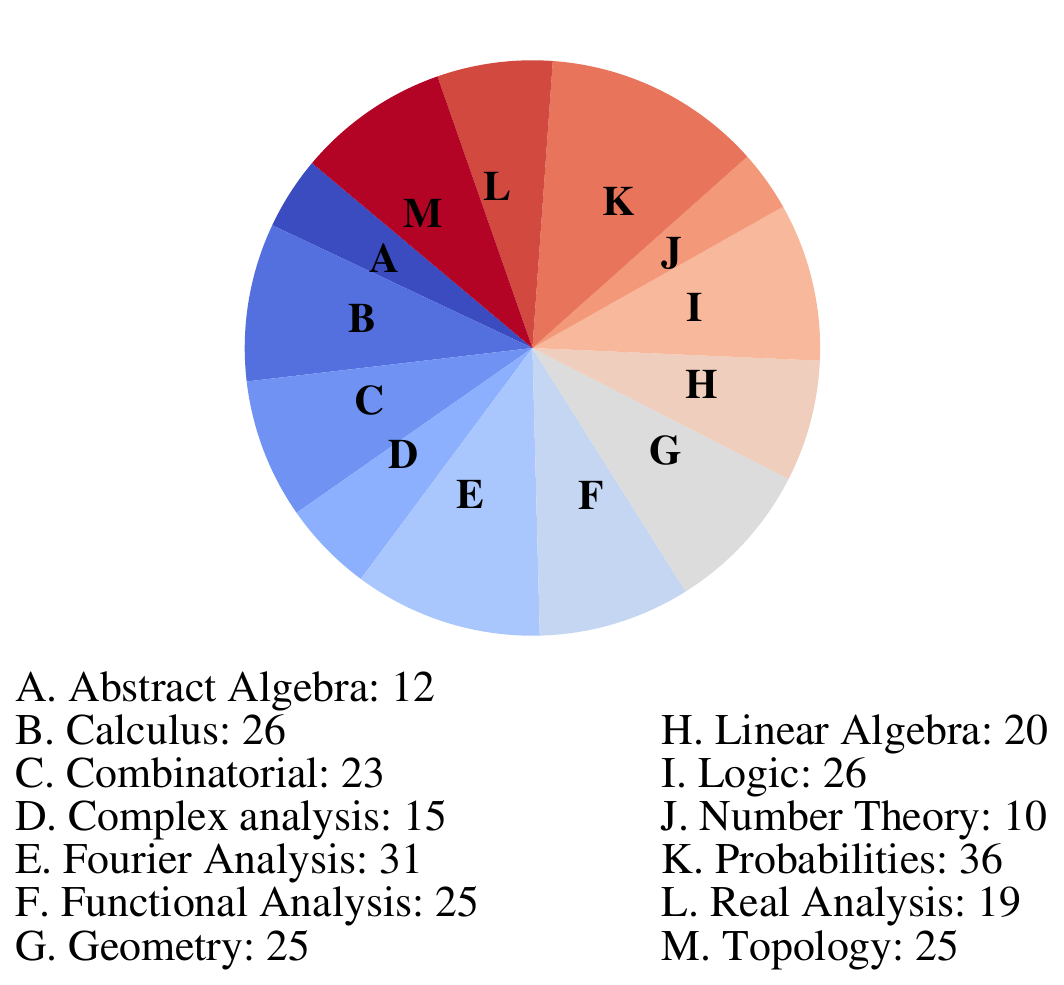}
    % \vspace{-2em}
    \caption{Multiple-choice Reasoning Problems by Category (Total = 293).}
    \label{fig:mc_by_category}
\end{figure}

\begin{figure}[h!]
    \centering
    \vspace{-0.5em}\includegraphics[width=0.65\linewidth]{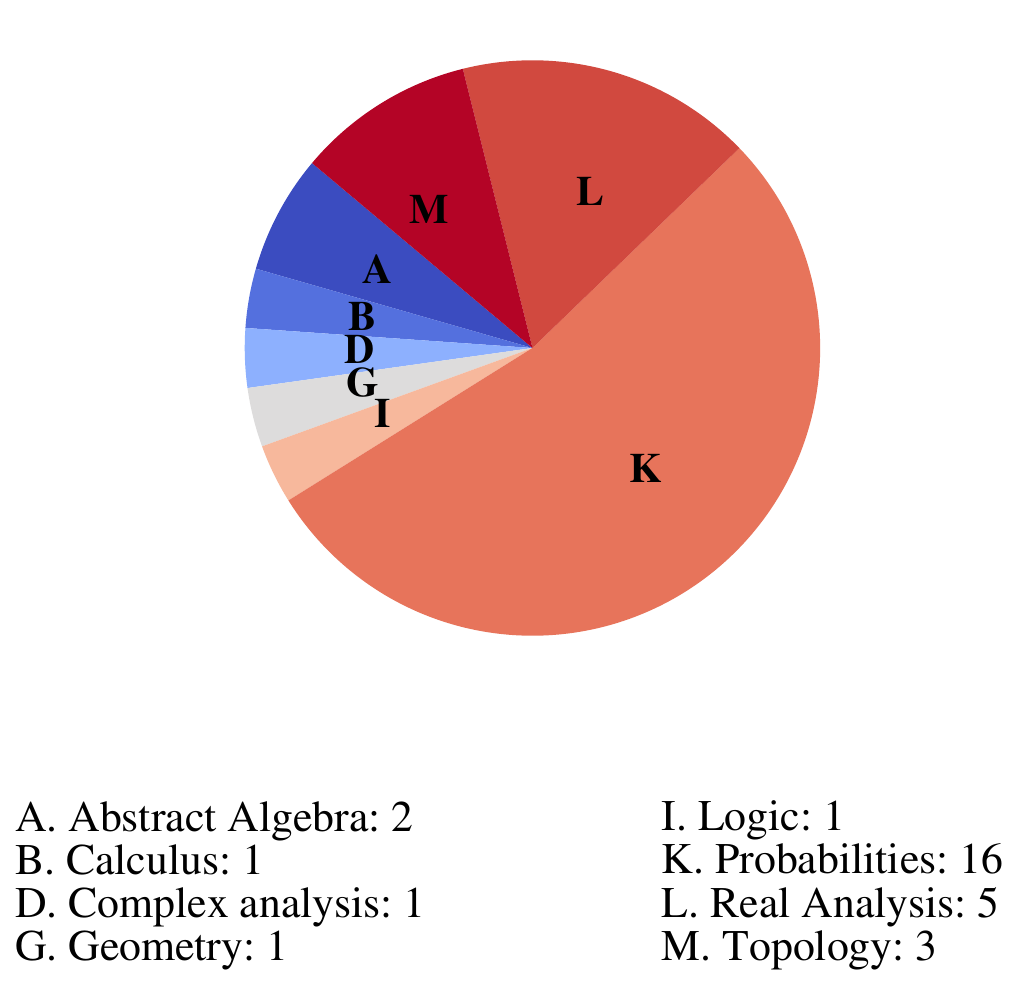}
    % \vspace{-3em}
    \caption{Transformed Problems by Category (Total = 30).}
    \label{fig:transformed_by_category}
\end{figure}

\begin{figure}[h!]
    \centering
    \vspace{-2em}\includegraphics[width=0.65\linewidth]{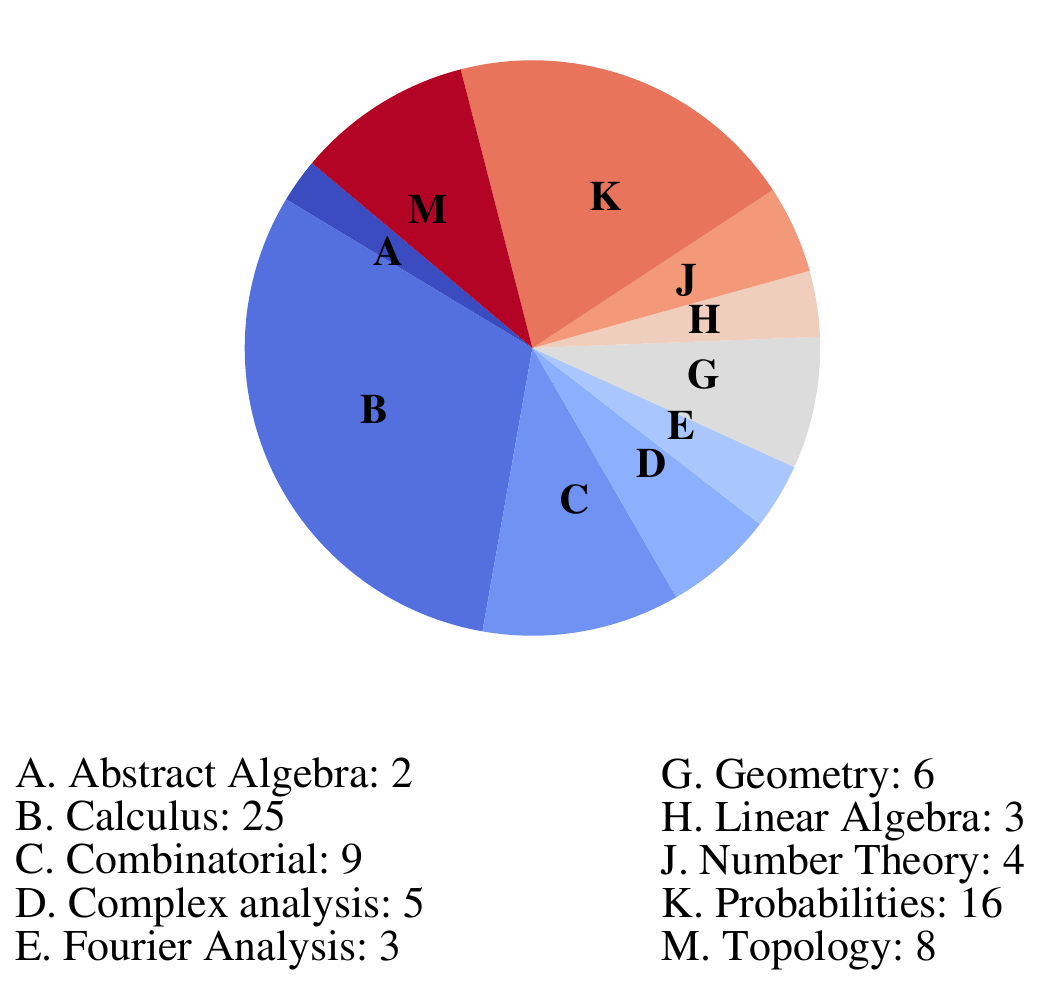}
    % \vspace{-3em}
    \caption{direct-answer Problems by Category (Total = 81).}\label{fig:computational_by_category}
\end{figure}

\section{Data sources}
\label{app:data_sources}
The following is a complete list of sources \benchname draws from:

\begin{itemize}
    \item Probability:
\cite{gamarnik_stochastic_2013,grinstead_probability_2006,lehman_mcs_2017}

\item Geometry / Topology:
\cite{hatcher_topology_2002,lee_manifolds_2013,petersen_riemannian_2006}

\item Abstract Algebra:
\cite{gallian_abstract_2021,coya_blog_2014}

\item Logic:
\cite{enderton_logic_2001,marker_modeltheory_2006,leary_logic_2015}

\item Real Analysis:
\cite{lebl_basic_analysis,royden_real_analysis}

\item Number Theory:
\cite{stein_number_theory,milne_ant,poonen_number_theory}

\item Complex Analysis:
\cite{cain_complex,lebl_complex,brown_complex}

\item Linear Algebra:
\cite{mit_linear_algebra,berkeley_eecs16b,hoell_ftla}

\item Fourier Analysis:
\cite{brown_fourier,stein_fourier,kammler_fourier}

\item Functional Analysis:
\cite{rudin_functional,reed_simon,buhler_functional}

\item Combinatorics:
\cite{bona_combinatorics,lovasz_combinatorics}

\item Calculus:
\cite{dawkins_calculus,strang_calculus}
\end{itemize}

\section{Autoformalization}
\label{app:autoformalization}

\textit{Initial Autoformalization}: For each problem, we employ an LLM-based autoformalizer to translate the natural language problem statement into Lean 4. The prompt configuration utilized for this step is the following: 
\texttt{\footnotesize You are an expert mathematician and Lean 4 formalizer. Your task is to translate the following informal mathematical statement and proof into standard Lean 4 code.
1. Ensure all syntax, imports, and tactics are strictly Lean 4. Do not output any Lean3 syntax.
2. Prioritize existing definitions, classes, and theorems from Mathlib. Do not define new terms or structures unless it is strictly necessary because the concept genuinely does not exist in the current library.
3. Before writing the code, explicitly state which specific Mathlib namespaces or files you are drawing from (e.g., Mathlib.ModelTheory.Basic, Mathlib.Combinatorics.SimpleGraph.Basic, or Mathlib.CategoryTheory.Logic).
4. Do not formalize the proof. Leave the proof blank except for a ``sorry'' statement. 
5. Format your answer as a valid Lean 4 file ready to be compiled. 
The Problem: [INSERT THEOREM OR DEFINITION HERE]}

\textit{Syntax Correction}: Standalone LLM formalizations remain highly unreliable for generating compilable Lean 4 theorems. To resolve initial syntax errors, the draft code is parsed by the Lean 4 compiler. If compilation fails, the resulting error messages are fed back into the autoformalizer in an iterative loop until the code compiles without syntax errors. The process is monitored by a human expert. 

\textit{Dual Semantic Verification}: To ensure the Lean 4 theorem statement accurately reflects the mathematical intent of the original natural language problem, we conduct semantic checking via a dual mechanism: an independent, secondary LLM evaluator and a human expert. 
Any semantic discrepancies feedback is routed back to the autoformalizer for correction.

\textit{Final Expert Review}: Once a statement passes both syntactic and initial semantic filters, an exhaustive final check will be conducted by another human expert before approving the formalization.

\subsection{Evaluating LLMs on Lean 4 Junk Theorems}
\label{app:junk-theorems}
In independent zero-shot conversations, models from multiple providers were presented with these snippets and asked two distinct questions:
\begin{enumerate}
    \item \textbf{Compilation:} Does this self-contained Lean 4 code compile? 
    \item \textbf{Semantics:} What does this code mean, and what theorem is it proving?
\end{enumerate}
A total of 360 calls (180 for compilation, 180 for semantics) were executed across four models: \texttt{gpt-5.5}, \texttt{gpt-4o}, \texttt{deepseek-v4-pro}, and \texttt{deepseek-chat}. 

\subsection{Results and Analysis}

Table~\ref{tab:junk_results} summarizes the performance of the models across both tasks. 

\begin{table*}[h]
\centering
\caption{Model performance on the Lean 4 Junk Theorems testbed.}
\label{tab:junk_results}
\begin{tabular}{@{}lcc@{}}
\toprule
\textbf{Model} & \textbf{Compilation Accuracy (\%)} & \textbf{Semantic Understanding (0--2)} \\ \midrule
\texttt{gpt-5.5} & 58\% & 2.00 \\
\texttt{deepseek-v4-pro} & 4\% & 1.91 \\
\texttt{deepseek-chat} & 4\% & 1.51 \\
\texttt{gpt-4o} & 18\% & 0.89 \\ \bottomrule
\end{tabular}
\end{table*}

\paragraph{Compilation Prediction:}
By construction, every snippet in the test corpus successfully compiles in Lean 4. However, the models heavily tended to predict the opposite. The misleading, ``nonsense'' mathematical reading of the junk theorems consistently overrode the models' ability to trace the actual type-checking logic. Accuracy ranged from a high of 58\% for \texttt{gpt-5.5} down to just 4\% for the DeepSeek models.

\paragraph{Semantic Meaning:}
The models' explanations of the theorems were hand-graded on a 0--2 scale based on whether the model could effectively ``see through'' the junk and explain the underlying formal trick (e.g., successfully recognizing the exploitation of junk values like \texttt{[].head! = 0} in empty lists). 

Here, the model rankings differed drastically from the compilation task. Most notably, \texttt{deepseek-v4-pro} demonstrated exceptional semantic understanding (scoring 1.91/2.00) while predicting compilation failures 96\% of the time. 

The experimental results highlight that understanding the trick of a junk theorem and trusting that it type-checks are fundamentally separate skills for current LLMs. Even when models accurately deduce the formal trick behind a statement, they frequently misjudge the boundary between plausible-looking Lean code and code that actually compiles. This finding motivates our design choice to use a verifier-in-the-loop framework, relying on both LLMs and human experts for semantic alignment while strictly delegating type-checking judgments to the Lean 4 compiler.

\subsection{Limitations of Current Mathlib Coverage}
\label{app:mathlib-coverage}

Consider the following candidate problem from our dataset, taken from \cite{petersen_riemannian_2006}:

\begin{quote}
\footnotesize
Let $(M, g)$ be a complete Riemannian manifold of dimension $n \geq 2$. Let $\mathrm{vol}$ denote the canonical Riemannian volume measure, and let $B(p, R)$ denote the open geodesic ball of radius $R$ centered at $p \in M$. Let $v(n, k, R)$ denote the volume of a geodesic ball of radius $R$ in the $n$-dimensional simply connected space form of constant sectional curvature $k$. Assume the Ricci curvature of $M$ satisfies the uniform lower bound $\mathrm{Ric} \geq (n-1)k$. If there exists a point $p \in M$ and a radius $R > 0$ (with $R \leq \pi/\sqrt{k}$ if $k > 0$) such that $\mathrm{vol}(B(p, R)) = v(n, k, R)$, then the metric $g$ has constant sectional curvature $k$ restricted to the ball $B(p, R)$.
\end{quote}

Even stating this theorem in Lean would require formal definitions of sectional curvature, Ricci curvature, geodesic balls, Riemannian volume, and model space forms. While Mathlib contains some foundations for smooth manifolds and Riemannian manifolds, many of the more specialized components of Riemannian geometry needed here are not yet available. Moreover, a standard proof of this result relies on the Bishop--Gromov inequality, a central theorem in comparison geometry that is also not formalized in Mathlib at the time of writing.

\section{Computational budgets}
\label{sec:app_comp_budget}
The computational budget for proof-step generation based models is usually computed with $K = N \times S \times T$ where $N$ is the number of independent attempts; $S$ is the number of tactics generated each time a node is expanded; and $T$ is the number of the expansion rounds (iterations) per run.
InternLM2-Math-Plus and InternLM2.5-Step-Prover both use the same search-style cap $(K = 1 \times 32 \times 100)$—one run per problem, 32 sampled continuations per expansion step, 100 expansion iterations, with early stop on first success. 
Gödel-Prover SFT and Gödel-Prover DPO each use pass@32. 
DeepSeek-Prover-V1.5 Base, SFT, and RL each use pass@128 parallel whole-proof sampling; DeepSeek-Prover-V1.5-RL with RMaxTS uses tree search with a default 128 LLM rollouts per problem plus upstream data repeat=16 for scheduling. 

The computational budget for whole-proof generation models is represented as pass@K by measuring the number of problems for which at least one valid proof is found among the top K generated attempts.
DeepSeek-R1-Distill each use pass@32 on Lean 4. 
MA-LoT’s two backbones share the same paper preset: 16 root proofs, 8 correction samples per proof, search depth 2 (documented as 16 + 8×2). 
Gödel-Prover-V2-32B uses pass@32 on the initial round plus self-correction: 2 correction rounds with 2 samples each after the initial 32 whole-proof draws.
DeepSeek-V3.2 uses pass@32 by default, DeepSeek-R1 uses pass@16. GPT-5.4 uses pass@32 and Claude-sonnet-5 uses pass@16. If applicable, the maximum token allowed per attempt is set to be 8192.

Aristotle, Codex and OpenGauss are agents rather than whole-proof generators, so each gets a single attempt per problem (pass@1), and its budget is limited by time and agent steps instead of a number of sampled proofs. 
Codex runs GPT-5.4 through the Codex CLI with one call per problem in a read-only sandbox and a 600 s timeout. OpenGauss runs GPT-5.4 with one agent query per problem; it can only read and edit files and is capped at 20 tool-calling turns and 300 s. Aristotle (Harmonic) is a hosted agent and we submit each problem once and wait up to 1800 s, and its internal compute budget is not visible to us.

\section{Evaluation process}
\label{app:evaluation_auxilary}
\subsection{direct-answer questions}
\label{app:evaluation_computational}

For answer generation, all models receive the same prompt, which instructs them to solve the problem and provide exactly one final answer enclosed in \texttt{boxed{...}}. Each problem is attempted up to five times using stochastic decoding with $T=0.7$ and $p=0.95$, with a maximum of 4096 generated tokens per attempt.
Answer extraction and grading follow a two-stage procedure. We first extract the expression enclosed in \texttt{boxed{...}} and check its equivalence to the reference answer using a procedure similar to that of Hendrycks et al.~\cite{hendrycks2021math}. If this stage fails but a boxed answer is successfully extracted, an LLM judge is used to determine whether the predicted answer is mathematically equivalent to the reference answer.

\subsection{Multiple-choice reasoning questions}
\label{app:evaluation_MC}

For answer generation, the system prompt instructs the model to answer only the multiple-choice question, using the theorem-proving problem as context, and to return a single option letter enclosed in parentheses (e.g., \texttt{(a)}). The theorem-proving problem and the corresponding MC question are concatenated into a single prompt. Each attempt is limited to 2048 generated tokens. Predictions are compared against the reference answers and evaluated using exact-match accuracy.

\section{Additional results}
\label{app:additiona_results}

In this section, we present additional results on model performance on \benchname. Table~\ref{tab:domain-interaction-accuracy} reports domain-level accuracy with and without interaction, while Tables~\ref{tab:deepseek-v15-base}--\ref{tab:codex-gpt5.4-domain} provide domain-level Lean 4 and multiple-choice results for individual models.

\begin{table}[t]
\centering
\resizebox{0.6\columnwidth}{!}{\scriptsize
\begin{tabular}{lcc}
\hline
 & DeepSeek-V3.2 & GPT-5.4 \\
\hline
Non-Interactive & 5.00\% & 9.06\% \\
Interactive & 7.50\% & 13.75\% \\
$\Delta$ & +2.50 pp & +4.69 pp \\
\hline
\end{tabular}}
\caption{Overall Lean 4 accuracy with and without interaction.}
\label{tab:compare_interactive}
\end{table}
\begin{table*}
\centering
\footnotesize
\caption{Lean 4 accuracy by domain with and without interaction.}
\label{tab:domain-interaction-accuracy}
\resizebox{\textwidth}{!}{%
\begin{tabular}{lrrrrrrr}
\hline
Domain & $N$ & DS Non-Int. & DS Int. & DS $\Delta$ & GPT Non-Int. & GPT Int. & GPT $\Delta$ \\
\hline
Abstract Algebra    & 30 & 10.0\% & 10.0\% & +0.0\% & 16.7\% & 23.3\% & +6.7\% \\
Calculus            & 26 & 3.9\%  & 3.9\%  & +0.0\% & 7.7\%  & 7.7\%  & +0.0\% \\
Combinatorial       & 23 & 4.3\%  & 4.3\%  & +0.0\% & 8.7\%  & 8.7\%  & +0.0\% \\
Complex Analysis    & 15 & 0.0\%  & 0.0\%  & +0.0\% & 0.0\%  & 0.0\%  & +0.0\% \\
Fourier Analysis    & 31 & 9.7\%  & 9.7\%  & +0.0\% & 9.7\%  & 12.9\% & +3.2\% \\
Functional Analysis & 25 & 4.0\%  & 4.0\%  & +0.0\% & 4.0\%  & 4.0\%  & +0.0\% \\
Geometry            & 24 & 0.0\%  & 8.3\%  & +8.3\% & 0.0\%  & 8.3\%  & +8.3\% \\
Linear Algebra      & 20 & 10.0\% & 20.0\% & +10.0\% & 35.0\% & 35.0\% & +0.0\% \\
Logic               & 23 & 0.0\%  & 8.7\%  & +8.7\% & 4.3\%  & 17.4\% & +13.0\% \\
Number Theory       & 10 & 20.0\% & 30.0\% & +10.0\% & 30.0\% & 40.0\% & +10.0\% \\
Probabilities       & 54 & 3.7\%  & 3.7\%  & +0.0\% & 3.7\%  & 9.3\%  & +5.6\% \\
Real Analysis       & 31 & 3.2\%  & 6.5\%  & +3.2\% & 9.7\%  & 19.4\% & +9.7\% \\
Topology            & 8  & 0.0\%  & 0.0\%  & +0.0\% & 0.0\%  & 0.0\%  & +0.0\% \\
\hline
\end{tabular}%
}
\end{table*}

\begin{table}[htbp]
\centering
\footnotesize
\caption{Accuracy by domain for DeepSeek V1.5 Base.}
\label{tab:deepseek-v15-base}
\begin{tabular}{lcc}
\hline
Domain & Lean 4 Acc. (\%) & MC Acc. (\%) \\
\hline
Abstract Algebra & 0.0 & 58.3 \\
Calculus & 0.0 & 48.1 \\
Combinatorial & 0.0 & 65.2 \\
Complex Analysis & 0.0 & 43.8 \\
Fourier Analysis & 3.2 & 54.8 \\
Functional Analysis & 0.0 & 48.0 \\
Geometry & 0.0 & 69.2 \\
Linear Algebra & 10.0 & 45.0 \\
Logic & 0.0 & 44.4 \\
Number Theory & 0.0 & 50.0 \\
Probabilities & 0.0 & 51.9 \\
Real Analysis & 3.2 & 70.0 \\
Topology & 0.0 & 57.1 \\
Overall & 1.2 & 54.3 \\
\hline
\end{tabular}
\end{table}
\begin{table}[htbp]
\centering
\footnotesize
\caption{Accuracy by domain for DeepSeek V1.5 SFT.}
\label{tab:deepseek-v15-sft}
\begin{tabular}{lcc}
\hline
Domain & Lean 4 Acc. (\%) & MC Acc. (\%) \\
\hline
Abstract Algebra & 16.7 & 16.7 \\
Calculus & 0.0 & 37.0 \\
Combinatorial & 4.3 & 43.5 \\
Complex Analysis & 0.0 & 37.5 \\
Fourier Analysis & 9.7 & 29.0 \\
Functional Analysis & 4.0 & 32.0 \\
Geometry & 8.3 & 38.5 \\
Linear Algebra & 20.0 & 30.0 \\
Logic & 13.0 & 29.6 \\
Number Theory & 30.0 & 50.0 \\
Probabilities & 9.3 & 32.7 \\
Real Analysis & 6.5 & 35.0 \\
Topology & 0.0 & 21.4 \\
Overall & 9.1 & 32.8 \\
\hline
\end{tabular}
\end{table}
\begin{table}[htbp]
\centering
\footnotesize
\caption{Accuracy by domain for DeepSeek V1.5 RL.}
\label{tab:deepseek-v15-rl}
\begin{tabular}{lcc}
\hline
Domain & Lean 4 Acc. (\%) & MC Acc. (\%) \\
\hline
Abstract Algebra & 16.7 & 41.7 \\
Calculus & 7.7 & 18.5 \\
Combinatorial & 4.3 & 52.2 \\
Complex Analysis & 0.0 & 25.0 \\
Fourier Analysis & 9.7 & 22.6 \\
Functional Analysis & 4.0 & 44.0 \\
Geometry & 8.3 & 50.0 \\
Linear Algebra & 30.0 & 50.0 \\
Logic & 17.4 & 33.3 \\
Number Theory & 50.0 & 30.0 \\
Probabilities & 11.1 & 23.1 \\
Real Analysis & 3.2 & 35.0 \\
Topology & 0.0 & 35.7 \\
Overall & 11.2 & 34.1 \\
\hline
\end{tabular}
\end{table}
\begin{table}[htbp]
\centering
\footnotesize
\caption{Accuracy by domain for RMaxTS.}
\label{tab:rmaxts}
\begin{tabular}{lcc}
\hline
Domain & Lean 4 Acc. (\%) & MC Acc. (\%) \\
\hline
Abstract Algebra & 20.0 & 41.7 \\
Calculus & 15.4 & 29.6 \\
Combinatorial & 4.3 & 30.4 \\
Complex Analysis & 0.0 & 37.5 \\
Fourier Analysis & 12.9 & 48.4 \\
Functional Analysis & 4.0 & 44.0 \\
Geometry & 12.5 & 30.8 \\
Linear Algebra & 50.0 & 40.0 \\
Logic & 26.1 & 33.3 \\
Number Theory & 60.0 & 80.0 \\
Probabilities & 13.0 & 34.6 \\
Real Analysis & 16.1 & 40.0 \\
Topology & 0.0 & 39.3 \\
Overall & 16.6 & 38.5 \\
\hline
\end{tabular}
\end{table}

\begin{table}[htbp]
\centering
\footnotesize
\caption{Accuracy by domain for InternLM2-Math.}
\label{tab:internlm2-math}
\begin{tabular}{lcc}
\hline
Domain & Lean 4 Acc. (\%) & MC Acc. (\%) \\
\hline
Abstract Algebra & 23.3 & 75.0 \\
Calculus & 19.2 & 70.4 \\
Combinatorial & 4.3 & 73.9 \\
Complex Analysis & 0.0 & 68.8 \\
Fourier Analysis & 9.7 & 54.8 \\
Functional Analysis & 4.0 & 60.0 \\
Geometry & 8.3 & 88.5 \\
Linear Algebra & 30.0 & 60.0 \\
Logic & 13.0 & 55.6 \\
Number Theory & 40.0 & 40.0 \\
Probabilities & 9.3 & 42.3 \\
Real Analysis & 16.1 & 90.0 \\
Topology & 0.0 & 64.3 \\
Overall & 13.1 & 63.1 \\
\hline
\end{tabular}
\end{table}
\begin{table}[htbp]
\centering
\footnotesize
\caption{Accuracy by domain for InternLM2.5 Step Prover.}
\label{tab:internlm2-5}
\begin{tabular}{lccc}
\hline
Domain & Solved & Total & Lean 4 Acc. (\%) \\
\hline
Abstract Algebra & 5 & 30 & 16.7 \\
Calculus & 4 & 26 & 15.4 \\
Combinatorial & 0 & 23 & 0.0 \\
Complex Analysis & 0 & 15 & 0.0 \\
Fourier Analysis & 3 & 31 & 9.7 \\
Functional Analysis & 1 & 25 & 4.0 \\
Geometry & 2 & 24 & 8.3 \\
Linear Algebra & 4 & 20 & 20.0 \\
Logic & 2 & 23 & 8.7 \\
Number Theory & 3 & 10 & 30.0 \\
Probabilities & 6 & 54 & 11.1 \\
Real Analysis & 3 & 31 & 9.7 \\
Topology & 0 & 28 & 0.0 \\
Overall & 33 & 320 & 10.3 \\
\hline
\end{tabular}
\end{table}
\begin{table}[htbp]
\centering
\footnotesize
\caption{Accuracy by domain for Goedel SFT.}
\label{tab:goedel-sft}
\begin{tabular}{lcc}
\hline
Domain & Lean 4 Acc. (\%) & MC Acc. (\%) \\
\hline
Abstract Algebra & 16.7 & 16.7 \\
Calculus & 11.5 & 22.2 \\
Combinatorial & 4.3 & 26.1 \\
Complex Analysis & 0.0 & 18.8 \\
Fourier Analysis & 9.7 & 12.9 \\
Functional Analysis & 4.0 & 28.0 \\
Geometry & 4.2 & 26.9 \\
Linear Algebra & 30.0 & 5.0 \\
Logic & 13.0 & 18.5 \\
Number Theory & 40.0 & 10.0 \\
Probabilities & 11.1 & 25.0 \\
Real Analysis & 3.2 & 15.0 \\
Topology & 0.0 & 39.3 \\
Overall & 10.6 & 21.8 \\
\hline
\end{tabular}
\end{table}
\begin{table}[htbp]
\centering
\footnotesize
\caption{Accuracy by domain for Goedel DPO.}
\label{tab:goedel-dpo}
\begin{tabular}{lcc}
\hline
Domain & Lean 4 Acc. (\%) & MC Acc. (\%) \\
\hline
Abstract Algebra & 10.0 & 16.7 \\
Calculus & 15.4 & 22.2 \\
Combinatorial & 4.3 & 26.1 \\
Complex Analysis & 0.0 & 12.5 \\
Fourier Analysis & 9.7 & 16.1 \\
Functional Analysis & 8.0 & 28.0 \\
Geometry & 8.3 & 30.8 \\
Linear Algebra & 30.0 & 15.0 \\
Logic & 13.0 & 14.8 \\
Number Theory & 40.0 & 30.0 \\
Probabilities & 11.1 & 11.5 \\
Real Analysis & 6.5 & 15.0 \\
Topology & 0.0 & 17.9 \\
Overall & 11.2 & 18.9 \\
\hline
\end{tabular}
\end{table}
\begin{table}[htbp]
\centering
\footnotesize
\caption{Goedel v2 union outcomes aggregated by domain.}
\label{tab:goedelv2-union-outcomes-by-domain}
\begin{tabular}{lccc}
\hline
Domain & Solved & Total & Lean 4 Acc. (\%) \\
\hline
Abstract Algebra & 7 & 30 & 23.3 \\
Calculus & 5 & 26 & 19.2 \\
Combinatorial & 2 & 23 & 8.7 \\
Complex Analysis & 1 & 15 & 6.7 \\
Fourier Analysis & 4 & 31 & 12.9 \\
Functional Analysis & 1 & 25 & 4.0 \\
Geometry & 3 & 24 & 12.5 \\
Linear Algebra & 13 & 20 & 65.0 \\
Logic & 5 & 23 & 21.7 \\
Number Theory & 7 & 10 & 70.0 \\
Probabilities & 11 & 54 & 20.4 \\
Real Analysis & 11 & 31 & 35.5 \\
Topology & 0 & 8 & 0.0 \\
Overall & 70 & 320 & 21.9 \\
\hline
\end{tabular}
\end{table}
\begin{table}[htbp]
\centering
\footnotesize
\caption{Accuracy by domain for MA-LoT + DeepSeek V1.5.}
\label{tab:lot-dsv15}
\begin{tabular}{lccc}
\hline
Domain & Solved & Total & Lean 4 Acc. (\%) \\
\hline
Abstract Algebra & 4 & 30 & 13.3 \\
Calculus & 3 & 26 & 11.5 \\
Combinatorial & 1 & 23 & 4.3 \\
Complex Analysis & 3 & 15 & 20.0 \\
Fourier Analysis & 3 & 31 & 9.7 \\
Functional Analysis & 1 & 25 & 4.0 \\
Geometry & 8 & 24 & 33.3 \\
Linear Algebra & 4 & 20 & 20.0 \\
Logic & 4 & 23 & 17.4 \\
Number Theory & 5 & 10 & 50.0 \\
Probabilities & 5 & 54 & 9.3 \\
Real Analysis & 2 & 31 & 6.5 \\
Topology & 0 & 28 & 0.0 \\
Overall & 43 & 320 & 13.4 \\
\hline
\end{tabular}
\end{table}
\begin{table}[htbp]
\centering
\footnotesize
\caption{Accuracy by domain for MA-LoT + Goedel.}
\label{tab:lot-goedel}
\begin{tabular}{lccc}
\hline
Domain & Solved & Total & Lean 4 Acc. (\%) \\
\hline
Abstract Algebra & 3 & 30 & 10.0 \\
Calculus & 4 & 26 & 15.4 \\
Combinatorial & 1 & 23 & 4.3 \\
Complex Analysis & 2 & 15 & 13.3 \\
Fourier Analysis & 4 & 31 & 12.9 \\
Functional Analysis & 1 & 25 & 4.0 \\
Geometry & 8 & 24 & 33.3 \\
Linear Algebra & 7 & 20 & 35.0 \\
Logic & 3 & 23 & 13.0 \\
Number Theory & 5 & 10 & 50.0 \\
Probabilities & 6 & 54 & 11.1 \\
Real Analysis & 2 & 31 & 6.5 \\
Topology & 0 & 28 & 0.0 \\
Overall & 46 & 320 & 14.4 \\
\hline
\end{tabular}
\end{table}
\begin{table}[htbp]
\centering
\scriptsize
\caption{Accuracy by domain for GPT-5.4.}
\label{tab:gpt5.4-pass16}
\begin{tabular}{lccc}
\hline
Domain & Pass@16 (\%) & Pass@32 (\%) & MC Acc. (\%) \\
\hline
Abstract Algebra & 16.7 & 16.7 & 83.3 \\
Calculus & 7.7 & 7.7 & 77.8 \\
Combinatorial & 8.7 & 8.7 & 87.0 \\
Complex Analysis & 0.0 & 0.0 & 87.5 \\
Fourier Analysis & 9.7 & 9.7 & 71.0 \\
Functional Analysis & 4.0 & 4.0 & 84.0 \\
Geometry & 0.0 & 4.2 & 100.0 \\
Linear Algebra & 35.0 & 35.0 & 70.0 \\
Logic & 4.3 & 13.0 & 92.6 \\
Number Theory & 30.0 & 30.0 & 70.0 \\
Probabilities & 3.7 & 7.4 & 73.1 \\
Real Analysis & 9.7 & 9.7 & 90.0 \\
Topology & 0.0 & 0.0 & 85.7 \\
Overall & 9.1 & 10.6 & 82.0 \\
\hline
\end{tabular}
\end{table}
\begin{table}[htbp]
\centering
\footnotesize
\caption{Accuracy by domain for DeepSeek-R1.}
\label{tab:deepseek-reasoner-pass16}
\begin{tabular}{lcc}
\hline
Domain & Lean 4 Pass@16 (\%) & MC Acc. (\%) \\
\hline
Abstract Algebra & 10.0 & 75.0 \\
Calculus & 7.7 & 55.6 \\
Combinatorial & 4.3 & 65.2 \\
Complex Analysis & 26.7 & 62.5 \\
Fourier Analysis & 12.9 & 45.2 \\
Functional Analysis & 4.0 & 64.0 \\
Geometry & 33.3 & 73.1 \\
Linear Algebra & 35.0 & 20.0 \\
Logic & 26.1 & 55.6 \\
Number Theory & 40.0 & 50.0 \\
Probabilities & 11.1 & 40.4 \\
Real Analysis & 0.0 & 75.0 \\
Topology & 0.0 & 21.4 \\
Overall & 14.4 & 51.7 \\
\hline
\end{tabular}
\end{table}
\begin{table}[htbp]
\centering
\scriptsize
\caption{Accuracy by domain for DeepSeek-V3.2.}
\label{tab:deepseek-api-chat-pass32}
\begin{tabular}{lccc}
\hline
Domain & Pass@16 (\%) & Pass@32 (\%) & MC Acc. (\%) \\
\hline
Abstract Algebra & 10.0 & 10.0 & 66.7 \\
Calculus & 3.9 & 3.9 & 66.7 \\
Combinatorial & 4.3 & 4.3 & 87.0 \\
Complex Analysis & 0.0 & 0.0 & 75.0 \\
Fourier Analysis & 9.7 & 9.7 & 58.1 \\
Functional Analysis & 4.0 & 4.0 & 80.0 \\
Geometry & 0.0 & 0.0 & 88.5 \\
Linear Algebra & 10.0 & 10.0 & 80.0 \\
Logic & 0.0 & 0.0 & 70.4 \\
Number Theory & 20.0 & 20.0 & 70.0 \\
Probabilities & 3.7 & 5.6 & 59.6 \\
Real Analysis & 3.2 & 3.2 & 95.0 \\
Topology & 0.0 & 0.0 & 85.7 \\
Overall & 5.0 & 5.3 & 74.1 \\
\hline
\end{tabular}
\end{table}
\begin{table}[htbp]
\centering
\footnotesize
\caption{Lean 4 accuracy by domain for OpenGauss.}
\label{tab:opengauss-domain}
\begin{tabular}{lc}
\hline
Domain & Lean 4 Acc. (\%) \\
\hline
Abstract Algebra & 10.0 \\
Calculus & 3.8 \\
Combinatorics & 4.3 \\
Complex Analysis & 0.0 \\
Fourier Analysis & 9.7 \\
Functional Analysis & 4.0 \\
Geometry & 8.3 \\
Linear Algebra & 25.0 \\
Logic & 13.0 \\
Number Theory & 20.0 \\
Probability & 5.6 \\
Real Analysis & 6.5 \\
Topology & 0.0 \\
\hline
Overall & 8.1 \\
\hline
\end{tabular}
\end{table}
\begin{table}[htbp]
\centering
\footnotesize
\caption{Lean 4 and multiple-choice accuracy by domain for Codex (GPT-5.4).}
\label{tab:codex-gpt5.4-domain}
\begin{tabular}{lcc}
\hline
Domain & Lean 4 Acc. (\%) & MC Acc. (\%) \\
\hline
Abstract Algebra & 23.3 & 83.3 \\
Calculus & 7.7 & 81.5 \\
Combinatorics & 13.0 & 87.0 \\
Complex Analysis & 6.7 & 87.5 \\
Fourier Analysis & 9.7 & 71.0 \\
Functional Analysis & 8.0 & 88.0 \\
Geometry & 4.2 & 100.0 \\
Linear Algebra & 20.0 & 65.0 \\
Logic & 8.7 & 88.9 \\
Number Theory & 40.0 & 70.0 \\
Probability & 14.8 & 76.9 \\
Real Analysis & 25.8 & 90.0 \\
Topology & 0.0 & 100.0 \\
\hline
Overall & 14.1 & 83.9 \\
\hline
\end{tabular}
\end{table}

% \begin{table*}[htbp]
% \centering
% \footnotesize
% \caption{Original vs. transformed comparison across models for theorem proving in Lean and MC reasoning questions. Entries are counts only.}
% \label{tab:all-models-transformation-comparison}
% \begin{tabular}{lcccc}
% \hline
% Model & Lean 4 O$\checkmark$/T$\times$ & Lean 4 O$\times$/T$\checkmark$ & MC O$\checkmark$/T$\times$ & MC O$\times$/T$\checkmark$ \\
% \hline
% DeepSeek-Prover-V1.5 Base & 0 & 0 & 1 & 10 \\
% DeepSeek-Prover-V1.5 SFT & 2 & 2 & 3 & 2 \\
% DeepSeek-Prover-V1.5 RL & 4 & 2 & 3 & 5 \\
% DeepSeek-Prover-V1.5 RL + RMaxTS & 4 & 2 & 2 & 3 \\
% InternLM2-Math-Plus-7B & 2 & 2 & 1 & 2 \\
% InternLM2.5-StepProver & 3 & 2 & -- & -- \\
% DeepSeek-Prover + MA-LoT & 4 & 2 & -- & -- \\
% Goedel-Prover + MA-LoT & 4 & 2 & -- & -- \\
% Goedel-Prover-SFT & 4 & 2 & 5 & 6 \\
% Goedel-Prover-DPO & 4 & 2 & 2 & 0 \\
% Goedel-Prover-V2 & 6 & 0 & -- & -- \\
% GPT-5.4 & 2 & 1 & 2 & 0 \\
% DeepSeek-R1 & 4 & 0 & 2 & 0 \\
% DeepSeek-V3.2 & 3 & 1 & 2 & 1 \\
% \hline
% \end{tabular}
% \end{table*}

% \input{tables/computational_accuracy_by_category}
% \input{tables/computational_vs_lean4_overlap}

\clearpage

\section{Representative Error Examples}
\label{app:error_examples}
This appendix provides representative examples of the most common error categories observed during theorem proving. For each category, we include the original problem, a snippet of the generated Lean proof, the verifier feedback, and a brief analysis of the failure.

\begin{table*}[h]
\centering
\caption{Representative error categories.}
\begin{tabular}{ll}
\toprule
Category & Typical Cause \\
\midrule
Incorrect theorem application & Goal state mismatch \\
Incorrect tactic selection & Appropriate strategy, wrong tactic \\
Hallucinated Mathlib declarations & Non-existent theorem/library names \\
Parser and syntax errors & Invalid Lean syntax or proof context \\
Repeated-tactic loops & No progress despite continued generation \\
\bottomrule
\end{tabular}
\end{table*}

\subsection{Incorrect Theorem Application}
\label{app:error_example_incorrect_theorem}
\textbf{Problem.}
\textit{Q1: Let x be an element of a cyclic group of order 15. If exactly two of $x^3, x^5$, and $x^9$ are equal, prove that order of $x^{13} = 3$.}

\textbf{Model:} GPT-5.4

\textbf{Generated proof snippet:}
\begin{lstlisting}[basicstyle=\ttfamily\scriptsize]
have h15 : x ^ 15 = 1 := by
  rw [← hG]
  exact pow_card_eq_one x
have hx3 : x ^ 3 = 1 := by
  rcases h with h | h | h
  · rcases h with ⟨h35, _⟩
    have := congrArg (fun y => y * (x ^ (-3 : ℤ))) h35
    simp [zpow_natCast, mul_assoc] at this
    exact this
...
have horder_dvd : orderOf x ∣ 3 := by
  exact orderOf_dvd_of_pow_eq_one hx3
have horder_ne_one : orderOf x ≠ 1 := by
  intro ho
  apply hx_ne_one
  exact orderOf_eq_one_iff.mp ho
have horder_eq : orderOf x = 3 := by
  omega
simpa [horder_eq]
\end{lstlisting}

\textbf{Verifier feedback:}
\begin{lstlisting}[basicstyle=\ttfamily\scriptsize]
line 18, column 12-29:
function expected at
  pow_card_eq_one
term has type
  ?m.10070 ^ Fintype.card ?m.10067 = 1

line 15, column 28 to line 18, column 29:
unsolved goals
...
h15 : x ^ 15 = 1
|- orderOf (x ^ 13) = 3

line 19, column 4-8:
unexpected token 'have'; expected command
\end{lstlisting}

\textbf{Analysis.}
The model correctly identifies the high-level strategy of proving $x^3 = 1$ and rewriting $x^{13}=x$. However, in the branch $x^3=x^5$, it incorrectly derives $x^2=1$ rather than the required $x^3=1$. The resulting proof state no longer matches the intended argument, leaving the final goal unsolved.

\subsection{Incorrect Tactic Selection}
\label{app:error_example_incorrect_tactic}
\paragraph{Problem.}
\textit{Q1: Let x be an element of a cyclic group of order 15. If exactly two of $x^3, x^5$, and $x^9$ are equal, prove that order of $x^{13} = 3$.}

\paragraph{Model.}
GPT-5.4

\paragraph{Generated proof snippet.}
\begin{lstlisting}[basicstyle=\ttfamily\scriptsize]
have horder_dvd : orderOf x ∣ 3 := by
  exact orderOf_dvd_of_pow_eq_one hx3

have horder_ne_one : orderOf x ≠ 1 := by
  intro ho
  apply hx_ne_one
  exact orderOf_eq_one_iff.mp ho

have horder_eq : orderOf x = 3 := by
  omega
\end{lstlisting}

% \paragraph{Verifier feedback.}
% \begin{lstlisting}
% omega failed
% ...
% \end{lstlisting}

\paragraph{Analysis.}
The model selects a tactic that is inappropriate for the current goal: the tactic \texttt{omega} in the final step. The remaining goal involves divisibility rather than linear arithmetic. Although the overall proof direction is reasonable, the chosen tactic cannot establish the required divisibility property.

\subsection{Repeated-Tactic Loops}
\label{app:error_example_repeated}
\textbf{Problem.}
\textit{Q1: Let x be an element of a cyclic group of order 15. If exactly two of $x^3, x^5$, and $x^9$ are equal, prove that order of $x^{13} = 3$.}

\textbf{Model.}
DeepSeek-Prover-V1.5-RL

\textbf{Generated proof snippet.}
\begin{lstlisting}[basicstyle=\ttfamily\scriptsize]
have h₃ : x ^ 5 = x ^ 9 := by
  apply Eq.symm
  apply Eq.symm
  apply Eq.symm
  apply h₂
have h₄ : orderOf (x ^ 13) = 3 := by
  apply Eq.symm
  apply Eq.symm
  apply Eq.symm
  apply h₃
\end{lstlisting}

\textbf{Analysis.}
The model repeatedly applies the same tactic without making progress toward the goal. Such loops often consume the token budget while leaving the proof state essentially unchanged.

\section{Representative Success Pattern}

\subsection{Exact-lemma proofs}
\label{app:success_example_lemma}
\textbf{Problem.}
\textit{Q2: A Dedekind algebra is a triple $(N,s:N\to N, o\in N)$ s.t. $o\notin$ range($s$), $s$ is injective, and $N=$ clo$_s(o):=\cap \{X: o\in X$ and $(\forall x\in X)s(x)\in X\}$ (i.e. closure of all sets that are closed under $s$). Prove that for any Dedekind algebra $(N, s, o)$ and any formula $\phi(x)$ (possibly with parameters in $N)$, one has that if $\varphi(o)$ and $(\forall n\in N)(\varphi(n)\to \varphi(s(n))), $ then $(\forall n\in N)\varphi(n)$}

\textbf{Model.}
DeepSeek-R1

\textbf{Generated proof snippet.}
\begin{lstlisting}[basicstyle=\ttfamily\footnotesize]
intro n
have hclosure_set : sClosed A.s A.o {x | phi x} := by
  exact ⟨h0, λ x hx => hstep x hx⟩
exact (Set.mem_sInter.mp (A.carrier_is_closure n)) {x | phi x} hclosure_set
\end{lstlisting}

\textbf{Analysis.}
The model discharges its goal with \texttt{Set.mem\_sInter.mp} applied to the closure axiom.

\textbf{Problem.}
\textit{Q3: Prove that for any Von Neumann ordinals $\alpha,\beta,$ it holds that $\alpha \in \beta\vee \beta\in\alpha \vee \alpha = \beta$}

\textbf{Model.}
DeepSeek-R1

\textbf{Generated proof snippet.}
\begin{lstlisting}[basicstyle=\ttfamily\footnotesize]
intro h
rcases h with ⟨f, hsurj⟩
let D : Set ℕ := {n | n ∉ f n}
have hD : ∃ m, f m = D := hsurj D
rcases hD with ⟨m, hm⟩
have hmem : m ∈ D ↔ m ∉ f m := by
  simp [D]
have hmem' : m ∈ D ↔ m ∉ D := by
  rw [hm] at hmem
  exact hmem
by_cases h : m ∈ D
· exact (hmem'.mp h) h
· exact h (hmem'.mpr h)
\end{lstlisting}

\textbf{Analysis.}
The model reduces ordinal trichotomy to a single call of \texttt{lt\_trichotomy} followed by a case split.

\subsection{\texttt{have}-heavy proofs}
\label{app:success_example_have}
\textbf{Problem.}
\textit{Q4: Prove that $\mathcal{P}(\mathbb{N})$, the power set of $\mathbb{N}$, is not enumerable}

\textbf{Model.}
DeepSeek-R1

\textbf{Generated proof snippet.}
\begin{lstlisting}[basicstyle=\ttfamily\footnotesize]
intro h
rcases h with ⟨U, hUcomp, hU⟩
have hNone : Nat.Partrec (fun x : ℕ => Part.none) := Nat.Partrec.none
rcases hEnum.2 (fun x : ℕ => Part.none) hNone with ⟨i, hi⟩
have htot : ∀ x, phi i x = Part.some (U i x) := by
  intro x; exact hU i x
have hnone : ∀ x, phi i x = Part.none := by
  intro x; simpa [hi]
have hcontra : Part.some (U i 0) = Part.none := by
  calc
    Part.some (U i 0) = phi i 0 := by symm; exact htot 0
    _ = Part.none := hnone 0
exact (Part.some_ne_none (U i 0)) hcontra
\end{lstlisting}

\textbf{Problem.}
\textit{Q5: Assume the set of partial recursive unary functions is enumerable; that is, there is an enumeration $\{\varphi_i(x)\}_{i\in\omega}$ of partial recursive unary functions. Prove that there is no universal computable function; that is, the universal function $U:\mathbb{N}\times \mathbb{N} \to \mathbb{N}, U(k, x):= \phi_k(x)$ is not both total and computable}

\textbf{Model.}
DeepSeek-R1

\textbf{Generated proof snippet.}
\begin{lstlisting}[basicstyle=\ttfamily\footnotesize]
intro n
have hclosure_set : sClosed A.s A.o {x | phi x} := by
  exact ⟨h0, λ x hx => hstep x hx⟩
exact (Set.mem_sInter.mp (A.carrier_is_closure n)) {x | phi x} hclosure_set
\end{lstlisting}

\subsection{\texttt{have}-heavy proofs}
\label{app:success_example_trivial}
\textbf{Problem.}
\textit{Q6: Let $\mathcal{L} = \{c, f, A\}$ be a first order language 
with constant symbol $c$, a one-place (i.e. unary) function symbol $f$, and a two-place predicate (i.e. binary relation) symbol $A$. Let the structure $\cfM$ be defined by:
\begin{itemize}
    \item $|\cfM|$ = $\{1,2,3\}$
    \item $c^\cfM = 3$
    \item $f^{\cfM}(1) = 2, f^{\cfM}(2)=3, f^{\cfM}(3)=3$
    \item $A^\cfM = \{(1,2),(2,3),(3,3)\}$
\end{itemize}
Let $s(v)=1$ for all variables $v.$ For arbitrary variable $v$, a \emph{$v$-variant} of $s$ is an assignment $s[m/v]$ that disagrees with $s$ at most by assigning a different value $m$ to $v$. Show that $\cfM, s\models \exists x\Big(A(f(z),c)\to \forall y\big(A(y,x)\vee A(f(y),x)\big)\Big)$}

\textbf{Model.}
DeepSeek-R1

\textbf{Generated proof snippet.}
\begin{lstlisting}[basicstyle=\ttfamily\footnotesize]
refine ⟨M.three, λ h => ?_⟩
intro y
cases y
· right; exact True.intro
· left; exact True.intro
· left; exact True.intro
\end{lstlisting}

\section{Multi-turn prompt results example}
\subsection{More \texttt{have} proofs}
\label{app:multi-turn_have}
\textbf{Problem.}
\textit{Q7: Suppose 120 students take a final exam and the mean of their scores is 90. You have no other information about the students and the exam, that is, you should not assume that the highest possible score is 100. You may, however, assume that exam scores are nonnegative. Show that the best possible upper bound on the number of students who scored at least 180 is 60.}

\textbf{Model.}
GPT-5.4

\textbf{Successful proof snippet under interaction.}
\begin{lstlisting}[basicstyle=\ttfamily\footnotesize]
theorem problem_16
  (scores : Fin 120 → ℝ)
  (h_nonneg : ∀ i, 0 ≤ scores i)
  (h_mean : (∑ i, scores i) / 120 = 90) :
  (Finset.univ.filter (fun i => 180 ≤ scores i)).card ≤ 60 := by
  let S : Finset (Fin 120) := Finset.univ.filter (fun i => 180 ≤ scores i)
  have hsum : ∑ i, scores i = 10800 := by
    have h := h_mean
    norm_num at h
    linarith
  have hS_le_sum : (S.card : ℝ) * 180 ≤ ∑ i in S, scores i := by
    calc
      (S.card : ℝ) * 180 = ∑ i in S, 180 := by rw [Finset.sum_const, nsmul_eq_mul]
      _ ≤ ∑ i in S, scores i := by
        refine Finset.sum_le_sum ?_
        intro i hi
        exact Finset.mem_filter.mp hi |>.2
  have hsub_le_total : ∑ i in S, scores i ≤ ∑ i, scores i := by
    exact Finset.sum_le_univ_sum_of_nonneg (s := S) (f := scores) (by intro i; exact h_nonneg i)
  have hcard_le : (S.card : ℝ) * 180 ≤ 10800 := by
    linarith
  have hcard_real : (S.card : ℝ) ≤ 60 := by
    linarith
  exact_mod_cast hcard_real
\end{lstlisting}

\textbf{Failed proof snippet without interaction.}
\begin{lstlisting}[basicstyle=\ttfamily\footnotesize]
theorem problem_16
  (scores : Fin 120 → ℝ)
  (h_nonneg : ∀ i, 0 ≤ scores i)
  (h_mean : (∑ i, scores i) / 120 = 90) :
  (Finset.univ.filter (fun i => 180 ≤ scores i)).card ≤ 60 := by
  let S : Finset (Fin 120) := Finset.univ.filter (fun i => 180 ≤ scores i)
    have hsum : ∑ i, scores i = 10800 := by
      have h := h_mean
      norm_num at h ⊢
      linarith
    have h_le_sumS : 180 * S.card ≤ ∑ i in S, scores i := by
      calc
        180 * S.card = ∑ i in S, 180 := by rw [Finset.sum_const, nsmul_eq_mul]
        _ ≤ ∑ i in S, scores i := by
          refine Finset.sum_le_sum ?_
          intro i hi
          exact le_of_lt?
\end{lstlisting}

\textbf{Analysis.}
The model introduces intermediate \texttt{have} statements that are never used in the final proof, a pattern observed less frequently in the non-interactive setting.

\subsection{Manual longer proofs}
\label{app:multi-turn_manual}
\textbf{Problem.}
\textit{Q8: Let $\{X_i\}_{i\geq 1}$ be i.i.d. and $\mathbb{E}X_1>0$. Given $a>0$, show that $\mathbb{E}\tau<\infty$ for $\tau=\inf\{k\geq 1~:~S_k>a\}$, where $S_k=\sum_{i=1}^k X_i$.}

\textbf{Model.}
GPT-5.4

\textbf{Successful proof snippet under interaction.}
\begin{lstlisting}[basicstyle=\ttfamily\footnotesize]
theorem bona_1(n : ℕ) :
    Fintype.card { f : Fin n → Fin n // Function.Bijective f } = Nat.factorial n := by
  classical
  let e :
      { f : Fin n → Fin n // Function.Bijective f } ≃ Equiv.Perm (Fin n) :=
    { toFun := fun f =>
        Equiv.ofBijective f.1 f.2
      invFun := fun σ =>
        ⟨σ, σ.bijective⟩
      left_inv := by
        intro f
        apply Subtype.ext
        funext x
        rfl
      right_inv := by
        intro σ
        ext x
        rfl }
  calc
    Fintype.card { f : Fin n → Fin n // Function.Bijective f }
        = Fintype.card (Equiv.Perm (Fin n)) := Fintype.card_congr e
    _ = Nat.factorial (Fintype.card (Fin n)) := Fintype.card_perm
    _ = Nat.factorial n := by simp
\end{lstlisting}

\textbf{Failed proof snippet without interaction.}
\begin{lstlisting}[basicstyle=\ttfamily\footnotesize]
theorem bona_1(n : ℕ) :
    Fintype.card { f : Fin n → Fin n // Function.Bijective f } = Nat.factorial n := by
  simpa using Fintype.card_perm n
\end{lstlisting}

\textbf{Analysis.}
The proof under interactive mode transitions from \texttt{simpa using Fintype.card\_perm n} to an explicit \texttt{Equiv.Perm} construction through a \texttt{calc} chain.

\textbf{Problem.}
\textit{Q9: Prove that for any Von Neumann ordinals $\alpha,\beta,$ it holds that $\alpha \in \beta\vee$}

\textbf{Model.}
GPT-5.4

\textbf{Successful proof snippet under interaction.}
\begin{lstlisting}[basicstyle=\ttfamily\footnotesize]
theorem open_logic_9 (α β : Ordinal) :
  α < β ∨ β < α ∨ α = β := by
  by_cases h : α = β
  · exact Or.inr <| Or.inr h
  · have hne : α ≠ β := h
    rcases lt_or_gt_of_ne hne with hlt | hgt
    · exact Or.inl hlt
    · exact Or.inr <| Or.inl hgt
\end{lstlisting}

\textbf{Failed proof snippet without interaction.}
\begin{lstlisting}[basicstyle=\ttfamily\footnotesize]
theorem open_logic_9 (α β : Ordinal) :
  α < β ∨ β < α ∨ α = β := by
  simpa [lt_or_gt_of_ne, eq_comm] using lt_trichotomy α β
\end{lstlisting}

\textbf{Analysis.}
Under the interactive mode, the model replaces a \texttt{simpa}-with-\texttt{lt\_trichotomy} strategy with an explicit \texttt{by\_cases} argument using \texttt{lt\_or\_gt\_of\_ne}.

\end{document}